\documentclass[10pt,twocolumn,letterpaper]{article}
\usepackage[pagenumbers]{cvpr}

\usepackage{amsmath,amsfonts,bm}

\def\eqref#1{equation~\ref{#1}}

\def\1{\bm{1}}

\DeclareMathAlphabet{\mathsfit}{\encodingdefault}{\sfdefault}{m}{sl}
\SetMathAlphabet{\mathsfit}{bold}{\encodingdefault}{\sfdefault}{bx}{n}

\usepackage{array,colortbl,longtable}
\definecolor{groupgray}{rgb}{0.94,0.94,0.94}
\definecolor{oursblue}{rgb}{0.93,0.97,1.0}
\definecolor{gainGreen}{rgb}{0.05,0.50,0.18}
\definecolor{lossRed}{rgb}{0.80,0.10,0.10}
\newcommand{\gain}[1]{{\scriptsize\textcolor{gainGreen}{$\uparrow$#1}}}
\newcommand{\loss}[1]{{\scriptsize\textcolor{lossRed}{$\downarrow$#1}}}

\definecolor{cvprblue}{rgb}{0.21,0.49,0.74}
\usepackage[breaklinks,colorlinks,allcolors=cvprblue]{hyperref}
\def\confName{CVPR}
\def\confYear{2027}
\title{ImproveAnyTask: An Autonomous Post-Training Harness for Iterative Model Self-Improvement}
\author{%
  \small Xingbo Yao$^{1,3,*}$ \quad Xiaoman Wang$^{4,*}$ \quad Zhengwu Lei$^{3,5}$ \quad Tinghui Luo$^{3,5}$ \quad YiLin Zhang$^{3}$ \\
  \small Yuefeng Wu$^{3,6}$ \quad Tianfu Wang$^{1}$ \quad Qingyuan Zhan$^{3}$ \quad Ye Guo$^{3}$ \quad Daoxin Zhang$^{3}$ \\
  \small Zhe Xu$^{3}$ \quad Xuemin Zhao \quad Fengjie Zhu \quad Jian Liu$^{3,\dagger}$ \quad Hui Xiong$^{1,2,\dagger}$ \\[2pt]
  \small $^{1}$HKUST(GZ) \quad $^{2}$HKUST \quad $^{3}$Xiaohongshu Inc. \quad $^{4}$ECNU \quad $^{5}$ZJU \quad $^{6}$NTU \\
  \small $^{*}$Equal contribution. \quad $^{\dagger}$Corresponding authors.}

\makeatletter
\g@addto@macro\@maketitle{%
\par\noindent
\begin{minipage}{\textwidth}
  \captionsetup{hypcap=false}
  \centering
  \includegraphics[width=\textwidth]{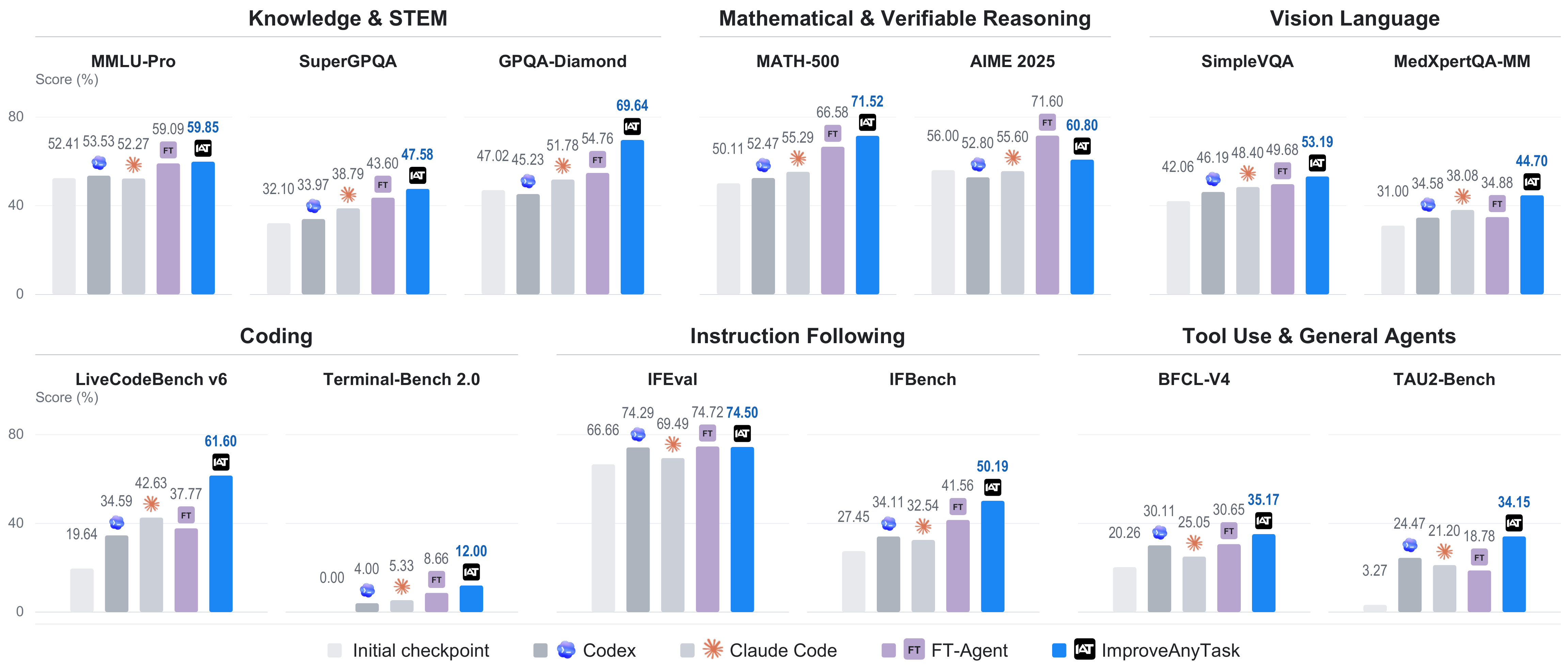}
  \captionof{figure}{\textbf{Task-specific improvement of Qwen3.5-4B-Base.}
  We compare the initial checkpoint with models produced by Codex, Claude Code, FT-Agent, and ImproveAnyTask across 13 tasks spanning six capability domains.
  Each run uses resources equivalent to eight H20 GPUs and stops after 24 hours or three optimization rounds, whichever is reached first.
  The mean gain is 17.45 percentage points, with a maximum of 41.96 points on LiveCodeBench v6. BFCL excludes Web Search and renormalizes the remaining weights.}
  \label{fig:teaser}
\end{minipage}
\par\vspace{6pt}
}
\makeatother
\begin{document}
\maketitle
\begin{abstract}
Adapting general-purpose large language models to specific tasks requires substantial human effort in designing data and training strategies. Sustaining improvement is especially challenging because model updates change the error distribution, making accurate diagnosis and strategy selection difficult. We introduce \textbf{ImproveAnyTask}, an autonomous post-training harness that improves task performance under a limited compute budget. Drawing inspiration from gradient-based parameter optimization, the harness organizes adaptation into error attribution, update-direction selection, and executable model updates. It combines metric-level and case-level analysis with targeted verification of uncertain causes, prioritizing problems by their measured recovery contribution. Research-backed strategies are compared for feasibility and assessed using gradient probes on the current checkpoint to estimate their potential to correct errors and their interference with existing capabilities. The selected strategy is translated into training data and a training configuration, with small-scale execution checks preceding full post-training. Subsequent evaluation guides model selection and further adaptation, while validated strategies and scripts are retained for reuse. Across 13 tasks, ImproveAnyTask achieves mean absolute gains of 17.45\% and 11.33\% on Qwen3.5-4B-Base and Qwen3.5-4B-Instruct, respectively, with a maximum gain of 41.96\% on LiveCodeBench v6, under a 24-hour budget with resources equivalent to eight H20 GPUs.
\end{abstract}

\section{Introduction}
\label{sec:introduction}

Recent large language models, including GLM-5, Qwen3.5, and Kimi K3, have demonstrated strong capabilities across diverse tasks~\citep{glm5,qwen35,kimi3}. However, strong general capabilities do not always translate into strong performance on specific tasks. Task-specific adaptation is therefore an important research direction for making these models more capable in real-world applications. Yet effective adaptation requires substantial effort in data curation, training strategy development, and model architecture design. Moreover, each model update can shift the error distribution, so sustained improvement requires continual refinement of the adaptation strategy.

Task-oriented post-training improves model performance through targeted data selection and student-aware curricula~\citep{rw_limo,rw_dmc,rw_dereason}, large-scale synthesis and distillation~\citep{rw_openthoughts,rw_opencodereasoning}, and structured supervision and learning objectives~\citep{rw_thinkrubrics,rw_generalreasoner}. These methods provide effective data and training recipes for specialized capabilities. However, selecting and adapting these recipes to a target task still \textit{requires substantial human expertise and time}, particularly as model updates change the remaining errors.

Recent systems automate post-training through iterative planning, strategy search, training, and evaluation~\citep{rw_ftdojo,rw_autotrainess,rw_lamdagent,rw_trex}. Complementary data-centric agents adapt data sources and curation pipelines using model feedback~\citep{rw_datamaster,rw_andes,rw_curateevo}. However, diagnosing model weaknesses and choosing improvement strategies remain largely dependent on the underlying LLM's interpretation of evaluation feedback. Errors in these decisions can lead to costly trial and error, making improvement \textit{inefficient and unreliable under a limited budget}. This motivates the following question:

\textit{Can a general-purpose LLM be improved on a target task autonomously, while making efficient use of a limited compute budget?}

To address this question, we propose \textbf{ImproveAnyTask}, an autonomous post-training harness that organizes task adaptation as an iterative model optimization process. Inspired by gradient-based parameter optimization, the harness connects error signals, update directions, and executable model updates. \textbf{Error Attribution} combines metric-level and case-level analysis, verifying uncertain causes with targeted inference tests before prioritizing a focal problem based on its measured recovery contribution to the task metric. \textbf{Update Direction} compares research-backed strategies in terms of applicability and reproduction difficulty, and probes their gradient alignment with desired corrections and existing capabilities on the current checkpoint to guide strategy selection before training. \textbf{Executable Model Updates} translates the selected strategy into training data and a training configuration, using small-scale execution and short training runs to identify implementation problems before full execution. Re-evaluation determines the outcome of each update and informs subsequent trials. Effective strategies and scripts are retained for reuse, while unsuccessful attempts provide evidence for revising future decisions. This process lets users adapt a model to a target task without manually redesigning data and training strategies at each iteration.

We evaluate 13 tasks spanning six capability domains using Qwen3.5-4B-Base and Qwen3.5-4B-Instruct~\citep{qwen35base,qwen35} under a 24-hour budget with resources equivalent to eight H20 GPUs. As shown in Figure~\ref{fig:teaser}, the Base-model gain averages 17.45\% and reaches 41.96\% on LiveCodeBench v6~\citep{livecodebench}. On some tasks, the optimized Base model even surpasses the Instruct model, whose manual post-training requires substantial time and resources. The Instruct-model gain averages 11.33\% (Table~\ref{tab:main-results}). These average gains exceed those of state-of-the-art baseline methods such as FT-Agent~\citep{rw_ftdojo} by absolute margins of 6.35\% on Base and 6.48\% on Instruct.

Our main contributions are:
\begin{itemize}
    \item We introduce \textbf{ImproveAnyTask}, a harness that organizes agents to autonomously improve LLM performance on a target task under a limited compute budget.
    \item To improve task performance \textbf{more efficiently}, we design several modules, including \textbf{Error Attribution} and \textbf{Update Direction}, to diagnose model errors more accurately and select suitable optimization directions across iterations.
    \item Experiments on 13 tasks show that ImproveAnyTask \textbf{improves} mean performance by \textbf{17.45\%} on Qwen3.5-4B-Base and \textbf{11.33\%} on Qwen3.5-4B-Instruct, with gains exceeding those of state-of-the-art baseline methods like FT-Agent~\citep{rw_ftdojo}.
\end{itemize}

\begin{figure*}[!t]
  \centering
  \includegraphics[width=\textwidth]{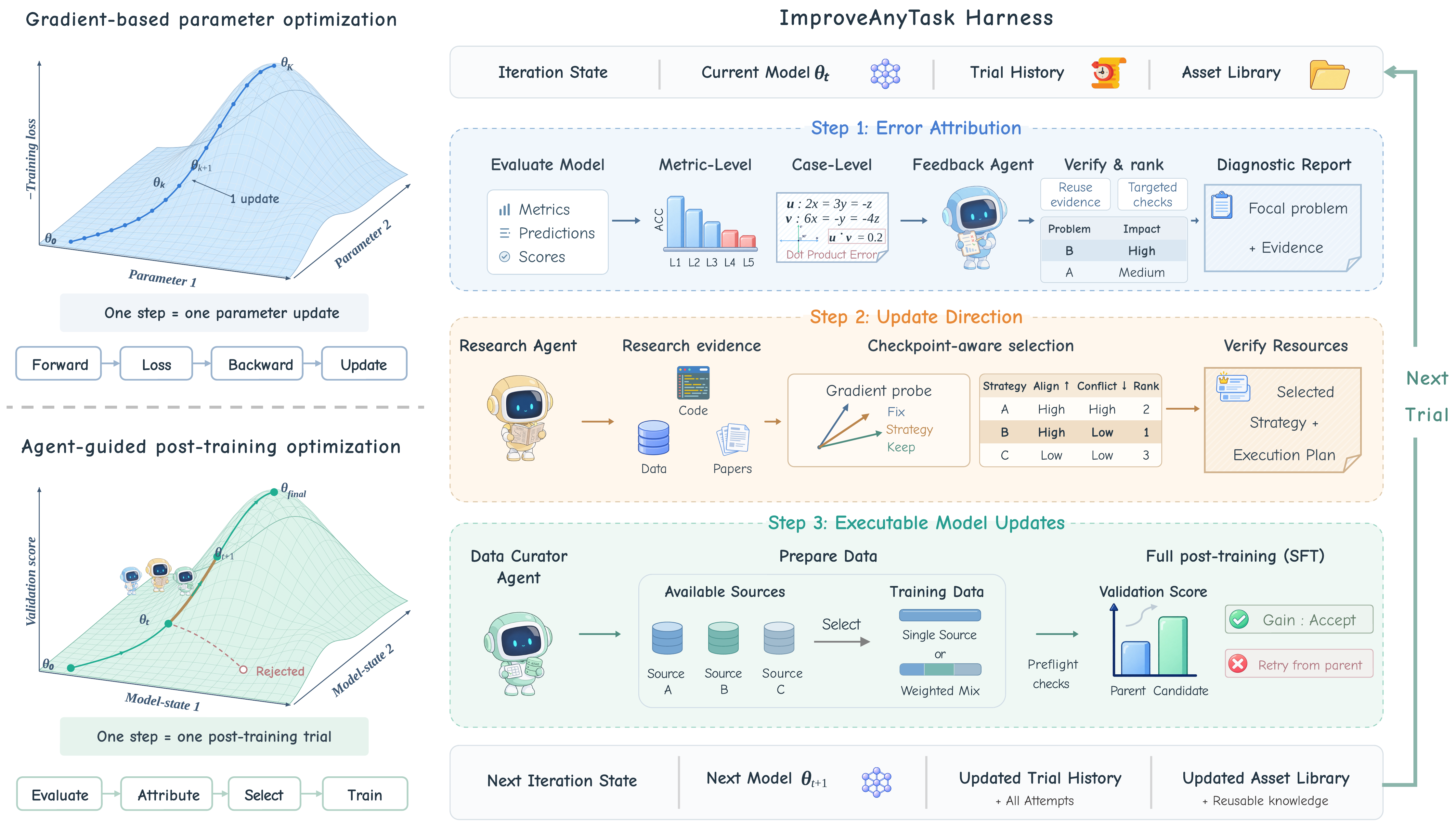}
  \caption{Overview of ImproveAnyTask. The harness treats task adaptation as iterative optimization: dual-level error analysis and targeted cause verification identify a focal problem; research evidence and checkpoint-specific gradient probes guide strategy selection; data preparation and post-training produce a candidate model; and re-evaluation determines the next checkpoint and records reusable assets.}
  \label{fig:framework}
\end{figure*}

\section{Related Work}
\label{sec:related-work}

\subsection{Task-Oriented Post-Training}
\label{sec:rw-task-post-training}

Recent work improves task performance by shaping both training data and learning signals. \textbf{Data selection and curricula} emphasize example quality, diversity, and compatibility with the learner. Compact reasoning datasets can elicit existing capabilities, while student-aware and stage-aware curricula adapt which examples are used and when~\citep{rw_limo,rw_naturalthoughts,rw_dmc,rw_dereason}. \textbf{Synthesis and distillation} expand supervision beyond existing labeled corpora: open data recipes scale reasoning and coding demonstrations~\citep{rw_openthoughts,rw_opencodereasoning}, iterative prompt synthesis improves the question supply~\citep{rw_promptcot}, and failure-driven methods target missing knowledge with document-grounded supervision~\citep{rw_logicsstem}. \textbf{Task-structured supervision} makes success criteria explicit through rubric-guided reasoning~\citep{rw_thinkrubrics}, executable software-repair tasks~\citep{rw_swesmith}, and diverse function-calling examples~\citep{rw_functiondiversity}. Complementarily, \textbf{training objectives and composition} determine how supervision translates into updates, including general-domain verification~\citep{rw_generalreasoner}, conditional-expectation rewards~\citep{rw_cer}, and multi-domain training or model merging~\citep{rw_m2rl}. These advances provide powerful optimization ingredients, and several already automate data construction or adapt it to the learner. 

However, selecting and revising these methods still requires \textbf{substantial human effort and time}. As training corrects some errors and introduces others, a previously effective strategy may no longer address the model's main weaknesses. Automated data construction alone does not resolve this need for \textbf{strategy adaptation as errors change}. ImproveAnyTask automates this repeated diagnosis and strategy revision, reducing the need for manual iteration.

\subsection{Autonomous Research and Model Improvement}
\label{sec:rw-autonomous-improvement}

LLM agents increasingly automate empirical research rather than merely propose ideas. AI Scientist-v2~\citep{rw_aiscientistv2} and AI-Researcher~\citep{rw_airesearcher} connect literature analysis, implementation, and experimentation, while MLE-STAR~\citep{rw_mlestar} improves machine-learning pipelines through search and targeted refinement. For model adaptation, PostTrainBench~\citep{rw_posttrainbench} evaluates autonomous post-training under bounded compute. FT-Agent~\citep{rw_ftdojo} and AutoTrainess~\citep{rw_autotrainess} combine iterative planning, error feedback, execution checks, and trial records. LaMDAgent~\citep{rw_lamdagent} searches end-to-end post-training pipelines, while TREX~\citep{rw_trex} combines literature research and execution with tree-based search and history reuse. LLMZero~\citep{rw_llmzero} and Pioneer Agent~\citep{rw_pioneer} further explore training trajectories and deployment-time interventions. These studies establish iterative, end-to-end post-training as a shared setting.

A complementary line makes training data an explicit optimization variable. DataMaster~\citep{rw_datamaster} searches external data and executable transformations, ANDES~\citep{rw_andes} provides adaptive instruction synthesis, and CurateEvo~\citep{rw_curateevo} evolves curation programs using post-training feedback. Curation-Bench~\citep{rw_curationbench} identifies an execution--research gap; its paper-adaptation scaffold requires agents to select prior methods, justify their applicability, and adapt them to available data and budgets. Related methods include SEAL~\citep{rw_seal}, which learns self-updates, and AgentEvolver~\citep{rw_agentevolver} and CoEvolve~\citep{rw_coevolve}, which evolve training tasks alongside the model. These approaches primarily optimize data strategies, with less emphasis on comparing candidate methods by expected benefit and reproduction difficulty. What is Missing from AI Post-Training AI~\citep{rw_missing_posttraining_ai} further finds that the evaluated post-training agents can repair execution problems but rarely reconsider their chosen strategies.

Although these works improve performance on specific tasks through autonomous post-training, they rely primarily on the LLM itself to diagnose the weaknesses of the current checkpoint and to determine the subsequent optimization direction. In practice, these decisions are\textbf{ frequently incorrect}, which wastes substantial time and can even drive optimization in the wrong direction.

\section{Methods}
\label{sec:methods}

\subsection{Problem Formulation and Overview}
\label{sec:problem-overview}

\noindent\textbf{Problem Formulation.}
Given an initial model $\theta_0$ and a target task $T$, our objective is to autonomously select and execute post-training strategies that improve its task performance within a time budget $\mathcal B$ of 24 hours, using resources equivalent to eight H20 GPUs:
\begin{equation}
\max_{\{s_t\}} J_T(\theta_{\mathrm{final}})
\quad \text{s.t.} \quad
T_{\mathrm{total}} \leq \mathcal B.
\label{eq:objective}
\end{equation}
Here, $J_T$ denotes the primary task metric, evaluated on the validation split during optimization; $s_t$ is the strategy adopted in trial $t$, and $\theta_{\mathrm{final}}$ is the selected post-trained candidate. $T_{\mathrm{total}}$ measures end-to-end elapsed time; Section~\ref{sec:experimental-setup} specifies the timing and stopping rules.

\noindent\textbf{Harness-Guided Model Optimization.}
As shown in Figure~\ref{fig:framework}, ImproveAnyTask draws inspiration from gradient-based parameter optimization. The harness first evaluates the current model $\theta_t$ on the validation split of $T$. The \textbf{Error Attribution module} then jointly analyzes metrics and failure cases and verifies uncertain causes through targeted inference tests. It produces a diagnostic report $F_t$ that identifies a focal problem $p_t$ and its supporting evidence. The \textbf{Update Direction module} combines research evidence, prior experience, and checkpoint-specific gradient probes to select a strategy $s_t$. The \textbf{Executable Model Updates module} translates $s_t$ into training data $D_t$ and a training configuration $h_t$, then performs post-training, producing a candidate model $\tilde{\theta}_{t+1}$. The candidate is subsequently re-evaluated.

\subsection{Error Attribution}
\label{sec:error-attribution}

\noindent\textbf{Challenge.} During error attribution, the agent often infers an incorrect cause or prioritizes a less consequential problem. Optimization guided by such a diagnosis wastes the limited budget and may even degrade existing capabilities. For example, in one run, the agent observed truncation in 4\% of the evaluated outputs and trained the model to produce shorter responses. However, this misdiagnosis led to lower accuracy after training.

\noindent\textbf{Dual-Level Error Attribution.} Metric-level analysis identifies weaknesses in overall performance and available subtask metrics. Case-level analysis examines individual predictions and scoring evidence to identify recurring failure patterns.

Combining both sources of evidence with available verification results, the harness guides the agent to select the focal problem from the candidate set $\mathcal{P}_t$:
\begin{equation}
  p_t = \arg\max_{p\in\mathcal{P}_t} I_T(p),
  \label{eq:focal-problem}
\end{equation}
where $I_T(p)$ denotes the assessed impact of problem $p$ on the primary metric. It represents a relative priority, not a numerical prediction of post-training gains.

\noindent\textbf{Verifying Candidate Causes.} For a problem $p$ with an uncertain cause, the agent formulates the cause as a falsifiable hypothesis and designs a test that requires no training. The harness executes the test and records the outcome. Behavioral causes arise from inference settings, such as the generation limit, prompt format, or answer parsing. They are tested by an inference-time intervention that removes the suspected cause before re-evaluating the failed examples. Capability gaps are tested by repeated sampling, which draws $k$ responses per failed example to test whether correct responses can be elicited. An example is counted as recovered only if it becomes correct under the primary metric, regardless of whether the symptom disappears.

The measured impact of $p$ is its recovered contribution to the primary metric:
\begin{equation}
  I_T(p) = \sum_{x\in E_p} w_x r_x,
  \label{eq:measured-impact}
\end{equation}
where $E_p$ contains failed examples attributed to $p$ with measured recovery outcomes, $w_x$ is the aggregation weight of example $x$ in the primary metric, and $r_x$ equals 1 if $x$ is recovered and 0 otherwise. Unknown recovery outcomes are excluded from the sum. Recovery provides diagnostic evidence, rather than proof of a unique cause or of post-training gains. Existing evidence is reused when it already verifies the cause. Additional tests are omitted when correcting all errors in a problem would still leave its contribution below the leading candidate.

Subsequent research and execution focus on $p_t$, rather than all identified problems. $F_t$ also stores the recovered outputs, which define the fix direction in Section~\ref{sec:update-direction}.

\subsection{Update Direction}
\label{sec:update-direction}

\noindent\textbf{Challenge.} A strategy that is effective in prior work may provide little new information to $\theta_t$. Applying it without regard to the current checkpoint may even disrupt the model's existing capability distribution.

\noindent\textbf{Structured Strategy Exploration.} The harness requires the agent to investigate multiple candidates for the same focal problem $p_t$ before selecting a strategy. Each candidate is recorded in a shared format covering its source, applicability, reported gains, reproduction difficulty, and reproduction steps.

The module then guides the agent to shortlist candidates for checkpoint-specific assessment. Before execution, the agent verifies the required resources and key reproduction steps of each shortlisted candidate.

\noindent\textbf{Outcome-Aware Reuse.} Each round considers newly researched strategies, previous candidates and trial outcomes, and validated assets. These records inform strategy selection but do not replace validation on the current model.

\noindent\textbf{Checkpoint-Aware Selection.} To obtain checkpoint-specific evidence, the harness provides the agent with a gradient probe based on first-order influence estimation~\citep{pruthi2020tracin,xia2024less}. For each candidate strategy $s$, the agent prepares a small set of training examples. On the fixed $\theta_t$, the probe computes projected per-example loss gradients and aggregates them into three directions. The fix direction $g_{\mathrm{fix}}$ averages the gradients of the recovered outputs of $p_t$ and represents the behavior the model should acquire. The keep direction $g^{(j)}_{\mathrm{keep}}$ averages the gradients of correctly answered examples in subcategory $j$ and represents the capabilities the model should retain. The agent determines which subcategories to protect. The candidate direction $\bar g_s$ averages the gradients of the training examples of $s$.

When recovered outputs are unavailable, the probe uses reference solutions or teacher outputs verified by the task evaluator. Missing targets or zero-norm directions are reported as unavailable evidence rather than assigned a score.

Since a small gradient-descent step on $s$ reduces the loss of an example approximately in proportion to the inner product of their gradients, the probe reports the alignment score
\begin{equation}
  \begin{aligned}
    S(s) ={}& \cos\bigl(\bar g_s, g_{\mathrm{fix}}\bigr) \\
      &{}+ \beta \min_{j}\,\min\bigl(0,\cos(\bar g_s,g^{(j)}_{\mathrm{keep}})\bigr),
  \end{aligned}
  \label{eq:strategy-score}
\end{equation}
where $\cos(\cdot,\cdot)$ denotes cosine similarity and $\beta>0$ is a penalty weight. The first term measures alignment with the desired corrections, and the second penalizes the most severe conflict with a protected subcategory, as in gradient-constrained continual learning~\citep{lopezpaz2017gem}. Candidates are ranked by $S(s)$. The agent treats this score as evidence rather than a decision rule. It selects $s_t$ accordingly and records its rationale when choosing a lower-ranked candidate. If all candidates have valid scores and no candidate obtains a positive score, the agent returns to error attribution and proceeds to the problem with the next largest impact without spending a training trial.

\noindent\textbf{Risk and Step Size.} Correctly answered examples whose gradients have a negative inner product with $\bar g_{s_t}$ indicate potential interference under a small update. The probe reports these examples by subcategory, and the agent decides whether to add replay data drawn from examples the model already answers correctly. The probe also estimates how sensitive the model's outputs on correct examples are to an update along $\bar g_{s_t}$. Following the trust-region principle~\citep{schulman2015trpo}, the agent reduces the learning rate reported for $s_t$ when this sensitivity is high. These computations require no parameter update and complete in minutes, whereas a training trial takes hours.

\subsection{Executable Model Updates}
\label{sec:executable-updates}

The selected strategy $s_t$ is implemented as training data $D_t$ and a configuration $h_t$. Training settings follow the selected method, with batch size and parallelism adapted to the current model $\theta_t$ and the available resources. The harness then performs post-training:
\begin{equation}
\tilde{\theta}_{t+1} = \mathcal U(\theta_t; D_t, h_t),
\label{eq:model-update}
\end{equation}
where $\mathcal U$ denotes the training procedure and $\tilde{\theta}_{t+1}$ is the resulting candidate model.

\noindent\textbf{Execution-Guided Refinement.}
Before full execution, the agent applies the data preparation procedure to a small set of real samples. It checks whether their content, format, and training template match the requirements of $s_t$. A short training run verifies that the prepared samples can be loaded and parameter updates completed. When execution reveals errors or mismatches, the agent revises the processing tools, data pipeline, or $h_t$, then repeats the relevant checks. Full data preparation and training begin after these checks pass.

\noindent\textbf{Evaluation and Reuse.}
The candidate $\tilde{\theta}_{t+1}$ is re-evaluated on the same validation split and under the same conditions as $\theta_t$. Using $F_t$ and the executed strategy $s_t$, the agent analyzes changes in the primary metric, whether the focal problem $p_t$ has improved, and any new errors. If the candidate improves the validation score, it becomes $\theta_{t+1}$; otherwise, the next trial continues from the previous checkpoint, with $\theta_{t+1}=\theta_t$.

Experiment records link every attempt, including failures, to its strategy, data, training settings, and results; effective strategies, scripts, and data are retained in the asset libraries for reuse.

\section{Experiments}
\label{sec:experiments}

\subsection{Experimental Setup}
\label{sec:experimental-setup}

\noindent\textbf{Models.}
We evaluate ImproveAnyTask on \textbf{Qwen3.5-4B-Base} and \textbf{Qwen3.5-4B-Instruct}~\citep{qwen35base,qwen35}, testing task adaptation from pretrained and post-trained initializations. Each run starts independently from the original checkpoint.

\noindent\textbf{Tasks.}
Following the broad evaluation coverage used in recent frontier-model reports~\citep{qwen35,kimi3,glm5}, we evaluate task-specific self-improvement on 13 tasks across six capability domains: (1) \textbf{Knowledge and STEM}: MMLU-Pro, SuperGPQA, and GPQA-Diamond~\citep{mmlupro,supergpqa,gpqa}; (2) \textbf{Mathematical and Verifiable Reasoning}: MATH-500 and AIME 2025~\citep{math500,aime2025}; (3) \textbf{Instruction Following}: IFEval and IFBench~\citep{ifeval,ifbench}; (4) \textbf{Coding}: LiveCodeBench v6 and Terminal-Bench 2.0~\citep{livecodebench,terminalbench2}; (5) \textbf{Tool Use and General Agents}: BFCL-V4 and TAU2-Bench~\citep{bfcl,tau2bench}; and (6) \textbf{Vision Language}: SimpleVQA for short-answer visual factuality and MedXpertQA-MM for medical reasoning over images and clinical context~\citep{simplevqa,medxpertqa}. For BFCL-V4, we exclude the web-search subset because it requires an external paid search service.

\noindent\textbf{Compute Budget.}
Each run uses resources equivalent to \textbf{eight NVIDIA H20 GPUs} and stops after \textbf{24 hours} or \textbf{three optimization rounds}, whichever comes first. Timing spans the initial to final evaluation, including research, data preparation, validation, training, and failed attempts; unfinished work is interrupted at the deadline. Section~\ref{sec:budget-scaling} extends these limits.

\noindent\textbf{Evaluation Protocol.}
All methods share a fixed split of approximately 15\% validation and 85\% held-out test data (Appendix~\ref{app:evaluation-details}). Only validation data inform error analysis, strategy selection, and checkpoint selection; test data remain inaccessible during optimization. Within each run, we select the post-trained candidate with the highest validation score among those validated before the deadline; the initial checkpoint is excluded and reported separately as \textit{Original}.

All methods share the task-specific evaluation implementations and settings detailed in Appendix~\ref{app:evaluation-configuration}. To reduce sampling variability during optimization, we also set the agent's API temperature to $0$. Main and ablation results are averaged over two independent complete optimization runs per setting, each subject to the same budget. To reduce variability on the small AIME 2025 test set, we sample five responses per question with temperature $0.6$ and top-$p$ $0.95$, and report mean accuracy across responses and runs. We log exact matches to evaluation queries in training data. SimpleVQA uses Claude Opus 4.6 as the judge, with the same grading procedure for all methods; MedXpertQA-MM uses answer accuracy. BFCL excludes Web Search and renormalizes the remaining weights (Appendix~\ref{app:bfcl-scoring}).

\noindent\textbf{Baselines.}
We compare ImproveAnyTask against Codex~\citep{codex}, Claude Code~\citep{claudecode}, and FT-Agent~\citep{rw_ftdojo}. All systems use Opus 4.8, except Codex, which uses GPT-5.6-Sol and serves as a system-level baseline. For fair comparison, all systems use the same compute budget, tool permissions, and teacher-model access, with two independent runs on each of the 13 tasks. System configurations and reproduction instructions are provided in Appendix~\ref{app:implementation-details}. For training-data and teacher-answer generation, each agent autonomously selects Qwen3.5-27B or Qwen3.5-122B; all systems have access to the same teacher models.

\subsection{Main Results}
\label{sec:main-results}

\noindent\textbf{Overall Results.}
Table~\ref{tab:main-results} reports the main results. ImproveAnyTask achieves the highest overall scores of 51.91\% on Qwen3.5-4B-Base and 57.71\% on Qwen3.5-4B-Instruct, with mean gains of 17.45 and 11.33 percentage points over the initial models. FT-Agent is the strongest baseline on average, scoring 45.56\% and 51.23\%, respectively. ImproveAnyTask ranks first on 11 of 13 tasks for each initialization. FT-Agent performs better on Base AIME 2025 and IFEval, and on Instruct SuperGPQA and IFEval. These averages include the renormalized BFCL no-Web metric.

\noindent\textbf{Task-Level Results.}
ImproveAnyTask improves all thirteen Base-model tasks and twelve Instruct-model tasks. The largest Base-model gain occurs on LiveCodeBench v6, from 19.64\% to 61.60\% $(+41.96\%)$, while the smallest occurs on AIME 2025 $(+4.80\%)$. For Instruct, the largest gains are on LiveCodeBench v6 $(+29.12\%)$, AIME 2025 $(+23.60\%)$, and IFBench $(+21.56\%)$. Notably, under our evaluation settings, the optimized Base model outperforms the optimized Instruct model on MATH-500 and LiveCodeBench v6.

On IFEval, ImproveAnyTask gains 7.84 and 0.87 percentage points on Base and Instruct, respectively. Section~\ref{sec:budget-scaling} examines the Base-model trajectory within and beyond the first 24 hours.

SuperGPQA-Instruct is the only task with a decrease, from 49.78\% to 47.77\% $(-2.00\%)$, with changes computed before truncation. The recorded run contains three unsuccessful attempts. The first used 70K teacher-distilled examples, but only 12,747 were multiple-choice questions and the data were concentrated in physics, mathematics, and chemistry. The second used 15K student-consistent examples, of which 89.1\% reproduced responses the student already answered correctly. The third increased the correction ratio to 80\% using 1K targeted examples. These records suggest distribution mismatch and forgetting as possible explanations, but do not establish a single cause.

\noindent\textbf{Vision-Language Results.}
On SimpleVQA, ImproveAnyTask reaches 53.19\% on Base and 51.88\% on Instruct, improving over their initial checkpoints by 11.12 and 8.48 percentage points. On MedXpertQA-MM, it reaches 44.70\% and 50.41\%, with gains of 13.70 and 7.50 points. It outperforms all three baselines in all four visual-task comparisons, extending the observed benefits to general visual factuality and specialized medical question answering.

\begin{table*}[t]
\centering
\caption{Mean results over two independent optimization runs across 13 tasks; run-to-run variance is reported in the appendix. Each run selects its post-trained candidate by validation score, excluding Original. All scores are percentages; the change from Original is shown next to each adapted-model score. SimpleVQA is graded by Claude Opus 4.6.}
\label{tab:main-results}
\fontsize{8}{10}\selectfont
\setlength{\tabcolsep}{2.2pt}
\renewcommand{\arraystretch}{1.15}
\renewcommand{\gain}[1]{{\fontsize{6}{7}\selectfont\textcolor{gainGreen}{($\uparrow$#1)}}}
\renewcommand{\loss}[1]{{\fontsize{6}{7}\selectfont\textcolor{lossRed}{($\downarrow$#1)}}}
\newcommand{\resultcell}[2]{\mbox{#1\,#2}}
\resizebox{\linewidth}{!}{\begin{tabular}{@{}l@{\hspace{3pt}}cccc>{\columncolor{oursblue}}c@{\hspace{3pt}}cccc>{\columncolor{oursblue}[2.2pt][0pt]}c@{}}
\toprule
& \multicolumn{5}{c}{\textbf{Qwen3.5-4B-Base}}
& \multicolumn{5}{c}{\textbf{Qwen3.5-4B-Instruct}} \\
\cmidrule(lr){2-6}\cmidrule(lr){7-11}
\textbf{Benchmark}
& \textbf{Original} & \textbf{Codex} & \shortstack{\textbf{Claude}\\\textbf{Code}} & \textbf{FT-Agent} & \textbf{Ours}
& \textbf{Original} & \textbf{Codex} & \shortstack{\textbf{Claude}\\\textbf{Code}} & \textbf{FT-Agent} & \textbf{Ours} \\
\midrule
\rowcolor{groupgray}\multicolumn{11}{l}{\itshape Knowledge \& STEM} \\
MMLU-Pro
& 52.41
& \resultcell{53.53}{\gain{1.12}}
& \resultcell{52.27}{\loss{0.13}}
& \resultcell{59.09}{\gain{6.68}}
& \resultcell{\textbf{59.85}}{\gain{7.44}}
& 58.66
& \resultcell{53.41}{\loss{5.25}}
& \resultcell{57.27}{\loss{1.38}}
& \resultcell{65.26}{\gain{6.60}}
& \resultcell{\textbf{69.33}}{\gain{10.66}} \\

SuperGPQA
& 32.10
& \resultcell{33.97}{\gain{1.86}}
& \resultcell{38.79}{\gain{6.68}}
& \resultcell{43.60}{\gain{11.49}}
& \resultcell{\textbf{47.58}}{\gain{15.48}}
& 49.78
& \resultcell{49.54}{\loss{0.23}}
& \resultcell{47.04}{\loss{2.73}}
& \resultcell{\textbf{54.43}}{\gain{4.65}}
& \resultcell{47.77}{\loss{2.00}} \\

GPQA-Diamond
& 47.02
& \resultcell{45.23}{\loss{1.78}}
& \resultcell{51.78}{\gain{4.76}}
& \resultcell{54.76}{\gain{7.73}}
& \resultcell{\textbf{69.64}}{\gain{22.61}}
& 75.59
& \resultcell{73.21}{\loss{2.38}}
& \resultcell{78.57}{\gain{2.97}}
& \resultcell{69.64}{\loss{5.95}}
& \resultcell{\textbf{80.35}}{\gain{4.76}} \\

\midrule
\rowcolor{groupgray}\multicolumn{11}{l}{\itshape Mathematical \& Verifiable Reasoning} \\
MATH-500
& 50.11
& \resultcell{52.47}{\gain{2.35}}
& \resultcell{55.29}{\gain{5.17}}
& \resultcell{66.58}{\gain{16.47}}
& \resultcell{\textbf{71.52}}{\gain{21.41}}
& 51.29
& \resultcell{56.00}{\gain{4.70}}
& \resultcell{57.41}{\gain{6.11}}
& \resultcell{61.52}{\gain{10.23}}
& \resultcell{\textbf{63.52}}{\gain{12.23}} \\

AIME 2025
& 56.00
& \resultcell{52.80}{\loss{3.20}}
& \resultcell{55.60}{\loss{0.40}}
& \resultcell{\textbf{71.60}}{\gain{15.60}}
& \resultcell{60.80}{\gain{4.80}}
& 40.00
& \resultcell{44.40}{\gain{4.40}}
& \resultcell{56.40}{\gain{16.40}}
& \resultcell{50.80}{\gain{10.80}}
& \resultcell{\textbf{63.60}}{\gain{23.60}} \\

\midrule
\rowcolor{groupgray}\multicolumn{11}{l}{\itshape Instruction Following} \\
IFEval
& 66.66
& \resultcell{74.29}{\gain{7.62}}
& \resultcell{69.49}{\gain{2.83}}
& \resultcell{\textbf{74.72}}{\gain{8.06}}
& \resultcell{74.50}{\gain{7.84}}
& 82.78
& \resultcell{83.44}{\gain{0.65}}
& \resultcell{80.61}{\loss{2.17}}
& \resultcell{\textbf{85.18}}{\gain{2.39}}
& \resultcell{83.66}{\gain{0.87}} \\

IFBench
& 27.45
& \resultcell{34.11}{\gain{6.66}}
& \resultcell{32.54}{\gain{5.09}}
& \resultcell{41.56}{\gain{14.11}}
& \resultcell{\textbf{50.19}}{\gain{22.74}}
& 32.94
& \resultcell{35.68}{\gain{2.74}}
& \resultcell{36.07}{\gain{3.13}}
& \resultcell{42.54}{\gain{9.60}}
& \resultcell{\textbf{54.50}}{\gain{21.56}} \\

\midrule
\rowcolor{groupgray}\multicolumn{11}{l}{\itshape Coding} \\
LiveCodeBench v6
& 19.64
& \resultcell{34.59}{\gain{14.95}}
& \resultcell{42.63}{\gain{22.99}}
& \resultcell{37.77}{\gain{18.13}}
& \resultcell{\textbf{61.60}}{\gain{41.96}}
& 11.49
& \resultcell{24.77}{\gain{13.28}}
& \resultcell{22.43}{\gain{10.93}}
& \resultcell{17.96}{\gain{6.47}}
& \resultcell{\textbf{40.62}}{\gain{29.12}} \\

Terminal-Bench 2.0
& 0.00
& \resultcell{4.00}{\gain{4.00}}
& \resultcell{5.33}{\gain{5.33}}
& \resultcell{8.66}{\gain{8.66}}
& \resultcell{\textbf{12.00}}{\gain{12.00}}
& 10.66
& \resultcell{13.33}{\gain{2.66}}
& \resultcell{14.66}{\gain{4.00}}
& \resultcell{14.00}{\gain{3.33}}
& \resultcell{\textbf{21.33}}{\gain{10.66}} \\

\midrule
\rowcolor{groupgray}\multicolumn{11}{l}{\itshape Tool Use \& General Agents} \\
BFCL-V4$^\dagger$
& 20.26
& \resultcell{30.11}{\gain{9.85}}
& \resultcell{25.05}{\gain{4.78}}
& \resultcell{30.65}{\gain{10.38}}
& \resultcell{\textbf{35.17}}{\gain{14.91}}
& 66.73
& \resultcell{70.76}{\gain{4.02}}
& \resultcell{69.60}{\gain{2.86}}
& \resultcell{70.52}{\gain{3.78}}
& \resultcell{\textbf{78.33}}{\gain{11.60}} \\

TAU2-Bench
& 3.27
& \resultcell{24.47}{\gain{21.20}}
& \resultcell{21.20}{\gain{17.93}}
& \resultcell{18.78}{\gain{15.51}}
& \resultcell{\textbf{34.15}}{\gain{30.87}}
& 36.68
& \resultcell{39.21}{\gain{2.52}}
& \resultcell{40.25}{\gain{3.57}}
& \resultcell{43.89}{\gain{7.21}}
& \resultcell{\textbf{44.94}}{\gain{8.25}} \\

\midrule
\rowcolor{groupgray}\multicolumn{11}{l}{\itshape Vision Language} \\
SimpleVQA
& 42.06
& \resultcell{46.19}{\gain{4.12}}
& \resultcell{48.40}{\gain{6.33}}
& \resultcell{49.68}{\gain{7.61}}
& \resultcell{\textbf{53.19}}{\gain{11.12}}
& 43.40
& \resultcell{46.31}{\gain{2.90}}
& \resultcell{42.21}{\loss{1.19}}
& \resultcell{45.46}{\gain{2.06}}
& \resultcell{\textbf{51.88}}{\gain{8.48}} \\

MedXpertQA-MM
& 31.00
& \resultcell{34.58}{\gain{3.58}}
& \resultcell{38.08}{\gain{7.08}}
& \resultcell{34.88}{\gain{3.88}}
& \resultcell{\textbf{44.70}}{\gain{13.70}}
& 42.91
& \resultcell{41.50}{\loss{1.41}}
& \resultcell{46.20}{\gain{3.29}}
& \resultcell{44.76}{\gain{1.85}}
& \resultcell{\textbf{50.41}}{\gain{7.50}} \\

\midrule\rowcolor{oursblue}
\textbf{Overall Average}$^\ddagger$
& 34.46
& \resultcell{40.03}{\gain{5.57}}
& \resultcell{41.27}{\gain{6.81}}
& \resultcell{45.56}{\gain{11.10}}
& \resultcell{\textbf{51.91}}{\gain{17.45}}
& 46.38
& \resultcell{48.58}{\gain{2.20}}
& \resultcell{49.90}{\gain{3.52}}
& \resultcell{51.23}{\gain{4.85}}
& \resultcell{\textbf{57.71}}{\gain{11.33}} \\
\bottomrule
\end{tabular}}\par
\vspace{2pt}
{\scriptsize
$^\dagger$ BFCL excludes Web Search and renormalizes the remaining weights (Appendix~\ref{app:bfcl-scoring}).\par
$^\ddagger$ Columns average all 13 tasks. Bold marks the best result per model block; green/red denotes gains/losses.
}
\end{table*}

\subsection{Scaling the Optimization Budget}
\label{sec:budget-scaling}

\begin{figure*}[t]
    \centering
    \includegraphics[width=0.80\textwidth,trim=0 12bp 0 0,clip]{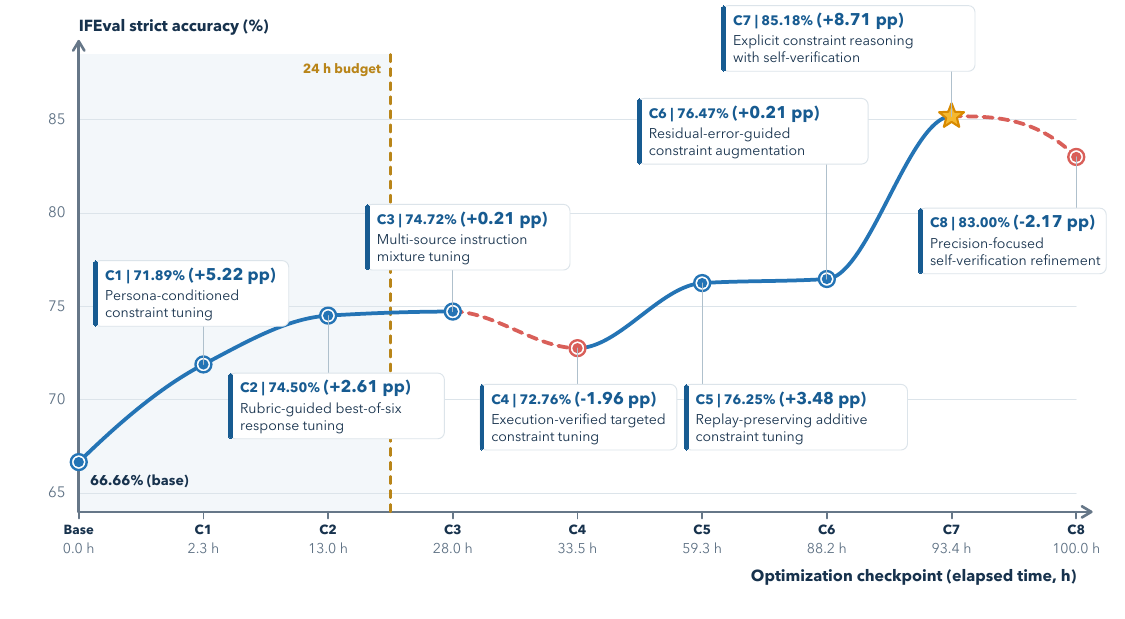}
    \caption{IFEval optimization over 100 hours. The dashed line marks 24 hours. Checkpoints are equally spaced; changes compare adjacent displayed checkpoints. C5 resumes from C3.}
    \label{fig:ifeval-extended-budget}
\end{figure*}

We extend the IFEval Base experiment from 24 to 100 hours under the same resource allocation, timing rules, and generation settings as the main experiments. Removing the three-round limit allows eight candidate checkpoints. Figure~\ref{fig:ifeval-extended-budget} reports strict prompt-level accuracy; Appendix~\ref{app:ifeval-extended} provides execution details.

The first two checkpoints use persona-conditioned instruction tuning and rubric-guided response selection, increasing accuracy from 66.66\% to 74.50\% within 24 hours. Adding a third source of general instruction data yields a smaller gain, reaching 74.72\% at C3.

Targeted, execution-verified tuning reduces accuracy at C4. C5 restarts from C3 and adds targeted examples without replacing the broader mixture; C6 targets residual errors. C7 introduces constraint reasoning and self-verification in the training data, reaching 85.18\%, 10.67 points above C2. A narrower continuation focused on precision constraints at C8 falls to 83.00\%. Longer exploration finds a better strategy in this run, but does not guarantee improvement at every checkpoint.

\subsection{Ablation Study}
\label{sec:ablation}

\noindent\textbf{Experimental Design.}
We evaluate \textbf{Error Attribution (EA)}, \textbf{Update Direction (UD)}, and \textbf{Execution-Guided Refinement (ER)} within Executable Model Updates through module-level replacement experiments that preserve the complete optimization loop. Individual submechanisms may yield modest effects relative to variation in LLM-driven decisions, data generation, and training. With a fixed budget and limited independent runs, we assess each module's combined contribution.

\textit{w/o EA} replaces the entire Error Attribution module with an unstructured summary of evaluation feedback. The agent identifies the focal problem using its own judgment.

\textit{w/o UD} replaces the entire Update Direction module with ordinary agent-led research. The agent selects the next training strategy from the researched candidates using its own judgment.

\textit{w/o ER} removes Execution-Guided Refinement from Executable Model Updates. The agent proceeds to full data preparation and training, debugging execution issues as they arise.

All variants retain the underlying agent, tool permissions, and evaluation--training loop. These replacements change the harness procedures while allowing the agent to analyze feedback, research methods, and repair code. Appendix~\ref{app:ablation-details} specifies the replacements and recorded execution paths.

We evaluate Qwen3.5-4B-Instruct on six representative benchmarks, one per capability domain, including SimpleVQA, following Section~\ref{sec:experimental-setup}. Each setting uses two independent complete optimization runs. The variants may select different data and training configurations as their trajectories diverge. The comparisons measure each module's total effect under the same budget; they do not isolate the contribution of individual submechanisms.

\noindent\textbf{Overall Effects.}
Table~\ref{tab:ablation-results} reports a full-harness average of 58.99\%, a gain of 11.61 percentage points over the initial model. Replacing Error Attribution, Update Direction, or Execution-Guided Refinement reduces the final average by 7.12, 5.06, and 9.22 points, respectively. The drop without EA highlights the benefit of accurately diagnosing and prioritizing model errors. The drop without UD demonstrates the value of selecting strategies suited to the current checkpoint. Together, these results show that better error attribution and direction selection improve overall task performance. Appendix~\ref{app:ablation-details} describes the recorded strategies and execution issues.

On SimpleVQA, the full harness reaches 51.88\%, compared with 46.80\%, 48.92\%, and 45.09\% without EA, UD, and ER, respectively. The corresponding drops of 5.08, 2.96, and 6.79 percentage points extend the ablation evidence to visual factuality.

\noindent\textbf{Optimization Trajectories.}
Figure~\ref{fig:ablation-trajectories} in Appendix~\ref{app:ablation-details} illustrates the MMLU-Pro and MATH-500 trajectories.

\renewcommand{\gain}[1]{%
  {\scriptsize\textcolor{gainGreen}{($\uparrow$#1)}}%
}
\renewcommand{\loss}[1]{%
  {\scriptsize\textcolor{lossRed}{($\downarrow$#1)}}%
}

\begin{table}[t]
\centering
\caption{Ablation results (\%) for Qwen3.5-4B-Instruct, averaged over two independent optimization runs on six tasks.}
\label{tab:ablation-results}
\fontsize{8}{9.2}\selectfont
\setlength{\tabcolsep}{1.5pt}
\renewcommand{\arraystretch}{1.22}
\renewcommand{\gain}[1]{{\fontsize{6}{7}\selectfont\textcolor{gainGreen}{$\uparrow$#1}}}
\renewcommand{\loss}[1]{{\fontsize{6}{7}\selectfont\textcolor{lossRed}{$\downarrow$#1}}}
\newcommand{\ablationcell}[2]{\mbox{#1\hspace{0.25pt}\raisebox{-1.4pt}{#2}}}

\resizebox{\columnwidth}{!}{\begin{tabular}{@{}lcccc>{\columncolor{oursblue}[1.5pt][0pt]}c@{}}
\toprule
\textbf{Task}
& \textbf{Original}
& \textbf{w/o EA}
& \textbf{w/o UD}
& \textbf{w/o ER}
& \shortstack{\textbf{Full}} \\
\midrule

MMLU-Pro
& 58.66
& \ablationcell{61.20}{\gain{2.53}}
& \ablationcell{60.30}{\gain{1.63}}
& \ablationcell{52.51}{\loss{6.15}}
& \ablationcell{\textbf{69.33}}{\gain{10.66}} \\

MATH-500
& 51.29
& \ablationcell{56.00}{\gain{4.70}}
& \ablationcell{56.70}{\gain{5.41}}
& \ablationcell{54.11}{\gain{2.82}}
& \ablationcell{\textbf{63.52}}{\gain{12.23}} \\

LCB v6
& 11.49
& \ablationcell{25.16}{\gain{13.67}}
& \ablationcell{32.81}{\gain{21.32}}
& \ablationcell{27.95}{\gain{16.46}}
& \ablationcell{\textbf{40.62}}{\gain{29.12}} \\

TAU2
& 36.68
& \ablationcell{39.73}{\gain{3.05}}
& \ablationcell{41.66}{\gain{4.98}}
& \ablationcell{38.83}{\gain{2.15}}
& \ablationcell{\textbf{44.94}}{\gain{8.25}} \\

IFEval
& 82.78
& \ablationcell{82.35}{\loss{0.43}}
& \ablationcell{83.22}{\gain{0.43}}
& \ablationcell{80.17}{\loss{2.61}}
& \ablationcell{\textbf{83.66}}{\gain{0.87}} \\

SVQA
& 43.40
& \ablationcell{46.80}{\gain{3.39}}
& \ablationcell{48.92}{\gain{5.52}}
& \ablationcell{45.09}{\gain{1.68}}
& \ablationcell{\textbf{51.88}}{\gain{8.48}} \\

\midrule
\rowcolor{oursblue}
\textbf{Average}
& 47.38
& \ablationcell{51.87}{\gain{4.49}}
& \ablationcell{53.94}{\gain{6.55}}
& \ablationcell{49.78}{\gain{2.39}}
& \ablationcell{\textbf{58.99}}{\gain{11.61}} \\

\bottomrule
\end{tabular}}\par

\vspace{2pt}

{\fontsize{7}{8}\selectfont
\parbox{\columnwidth}{\raggedright
LCB v6: LiveCodeBench v6; TAU2: TAU2-Bench; SVQA: SimpleVQA. Average covers all six tasks. Green/red values indicate changes from Original; bold marks the best result per row.
}}

\end{table}

\section{Conclusion}

We introduced ImproveAnyTask, an autonomous post-training harness that supports iterative model improvement through error attribution, research-grounded update selection, and executable model updates. Targeted verification and checkpoint-aware gradient probes ground diagnosis and strategy selection in evidence from the current model. Across 13 tasks, it achieves mean absolute gains of 17.45\% and 11.33\% on Qwen3.5-4B-Base and Qwen3.5-4B-Instruct under a 24-hour budget with resources equivalent to eight H20 GPUs. By automating repeated diagnosis, research, data preparation, and training, the harness reduces reliance on manual trial design and intervention. The results show that substantial task-specific gains can be achieved autonomously within a fixed budget, with the adapted Base model surpassing the original Instruct model on some tasks.

\section*{LLM Usage Statement}

The research presented in this paper, including the core ideas, experimental design, and quantitative results, is the original work of the authors. LLMs (e.g., ChatGPT) were used as writing assistants for tasks such as polishing prose, improving clarity, and correcting grammatical errors in the manuscript. All final content was reviewed and edited by the authors to ensure it accurately reflects our research and contributions.

{
\small
\bibliographystyle{ieeenat_fullname}
\bibliography{related_work_refs}

@misc{rw_limo,
      title={{LIMO: Less is More for Reasoning}}, 
      author={Yixin Ye and Zhen Huang and Yang Xiao and Ethan Chern and Shijie Xia and Pengfei Liu},
      year={2025},
      eprint={2502.03387},
      archivePrefix={arXiv},
      primaryClass={cs.CL},
      url={https://arxiv.org/abs/2502.03387}, 
}

@misc{rw_naturalthoughts,
      title={{NaturalThoughts: Selecting and Distilling Reasoning Traces for General Reasoning Tasks}}, 
      author={Yang Li and Youssef Emad and Karthik Padthe and Jack Lanchantin and Weizhe Yuan and Thao Nguyen and Jason Weston and Shang-Wen Li and Dong Wang and Ilia Kulikov and Xian Li},
      year={2025},
      eprint={2507.01921},
      archivePrefix={arXiv},
      primaryClass={cs.CL},
      url={https://arxiv.org/abs/2507.01921}, 
}

@misc{rw_dmc,
      title={{Tailoring the Curriculum: Student-Centered Reasoning Distillation via Dynamic Data-Model Compatibility}}, 
      author={Jiahao Huang and Fei Cheng and Junfeng Jiang and Akiko Aizawa},
      year={2026},
      eprint={2605.29229},
      archivePrefix={arXiv},
      primaryClass={cs.AI},
      url={https://arxiv.org/abs/2605.29229}, 
}

@misc{rw_dereason,
      title={{DeReason: A Difficulty-Aware Curriculum Improves Decoupled SFT-then-RL Training for General Reasoning}}, 
      author={Hanxu Hu and Yuxuan Wang and Maggie Huan and Jannis Vamvas and Yinya Huang and Zhijiang Guo and Rico Sennrich},
      year={2026},
      eprint={2603.11193},
      archivePrefix={arXiv},
      primaryClass={cs.CL},
      url={https://arxiv.org/abs/2603.11193}, 
}

@misc{rw_openthoughts,
      title={{OpenThoughts: Data Recipes for Reasoning Models}}, 
      author={Etash Guha and Ryan Marten and Sedrick Keh and Negin Raoof and Georgios Smyrnis and Hritik Bansal and Marianna Nezhurina and Jean Mercat and Trung Vu and Zayne Sprague and Ashima Suvarna and Benjamin Feuer and Liangyu Chen and Zaid Khan and Eric Frankel and Sachin Grover and Caroline Choi and Niklas Muennighoff and Shiye Su and Wanjia Zhao and John Yang and Shreyas Pimpalgaonkar and Kartik Sharma and Charlie Cheng-Jie Ji and Yichuan Deng and Sarah Pratt and Vivek Ramanujan and Jon Saad-Falcon and Jeffrey Li and Achal Dave and Alon Albalak and Kushal Arora and Blake Wulfe and Chinmay Hegde and Greg Durrett and Sewoong Oh and Mohit Bansal and Saadia Gabriel and Aditya Grover and Kai-Wei Chang and Vaishaal Shankar and Aaron Gokaslan and Mike A. Merrill and Tatsunori Hashimoto and Yejin Choi and Jenia Jitsev and Reinhard Heckel and Maheswaran Sathiamoorthy and Alexandros G. Dimakis and Ludwig Schmidt},
      year={2025},
      eprint={2506.04178},
      archivePrefix={arXiv},
      primaryClass={cs.LG},
      url={https://arxiv.org/abs/2506.04178}, 
}

@misc{rw_opencodereasoning,
      title={{OpenCodeReasoning: Advancing Data Distillation for Competitive Coding}}, 
      author={Wasi Uddin Ahmad and Sean Narenthiran and Somshubra Majumdar and Aleksander Ficek and Siddhartha Jain and Jocelyn Huang and Vahid Noroozi and Boris Ginsburg},
      year={2025},
      eprint={2504.01943},
      archivePrefix={arXiv},
      primaryClass={cs.CL},
      url={https://arxiv.org/abs/2504.01943}, 
}

@misc{rw_promptcot,
      title={{PromptCoT 2.0: Scaling Prompt Synthesis for Large Language Model Reasoning}}, 
      author={Xueliang Zhao and Wei Wu and Jian Guan and Zhuocheng Gong and Lingpeng Kong},
      year={2025},
      eprint={2509.19894},
      archivePrefix={arXiv},
      primaryClass={cs.LG},
      url={https://arxiv.org/abs/2509.19894}, 
}

@misc{rw_logicsstem,
      title={{Logics-STEM: Empowering LLM Reasoning via Failure-Driven Post-Training and Document Knowledge Enhancement}}, 
      author={Mingyu Xu and Cheng Fang and Keyue Jiang and Yuqian Zheng and Yanghua Xiao and Baojian Zhou and Qifang Zhao and Suhang Zheng and Xiuwen Zhu and Jiyang Tang and Yongchi Zhao and Yijia Luo and Zhiqi Bai and Yuchi Xu and Wenbo Su and Wei Wang and Bing Zhao and Lin Qu and Xiaoxiao Xu},
      year={2026},
      eprint={2601.01562},
      archivePrefix={arXiv},
      primaryClass={cs.AI},
      url={https://arxiv.org/abs/2601.01562}, 
}

@misc{rw_thinkrubrics,
      title={{Think-with-Rubrics: From External Evaluator to Internal Reasoning Guidance}}, 
      author={Jiachen Yu and Zhihao Xu and Junjie Wang and Yujiu Yang},
      year={2026},
      eprint={2605.07461},
      archivePrefix={arXiv},
      primaryClass={cs.CL},
      url={https://arxiv.org/abs/2605.07461}, 
}

@misc{rw_swesmith,
      title={{SWE-smith: Scaling Data for Software Engineering Agents}}, 
      author={John Yang and Kilian Lieret and Carlos E. Jimenez and Alexander Wettig and Kabir Khandpur and Yanzhe Zhang and Binyuan Hui and Ofir Press and Ludwig Schmidt and Diyi Yang},
      year={2025},
      eprint={2504.21798},
      archivePrefix={arXiv},
      primaryClass={cs.SE},
      url={https://arxiv.org/abs/2504.21798}, 
}

@misc{rw_functiondiversity,
      title={{Linguistic and Argument Diversity in Synthetic Data for Function-Calling Agents}}, 
      author={Dan Greenstein and Zohar Karnin and Chen Amiraz and Oren Somekh},
      year={2026},
      eprint={2601.17829},
      archivePrefix={arXiv},
      primaryClass={cs.CL},
      url={https://arxiv.org/abs/2601.17829}, 
}

@misc{rw_generalreasoner,
      title={{General-Reasoner: Advancing LLM Reasoning Across All Domains}}, 
      author={Xueguang Ma and Qian Liu and Dongfu Jiang and Ge Zhang and Zejun Ma and Wenhu Chen},
      year={2025},
      eprint={2505.14652},
      archivePrefix={arXiv},
      primaryClass={cs.CL},
      url={https://arxiv.org/abs/2505.14652}, 
}

@misc{rw_cer,
      title={{Reinforcement Learning with Conditional Expectation Reward}}, 
      author={Changyi Xiao and Caijun Xu and Yixin Cao},
      year={2026},
      eprint={2603.10624},
      archivePrefix={arXiv},
      primaryClass={cs.LG},
      url={https://arxiv.org/abs/2603.10624}, 
}

@misc{rw_m2rl,
      title={{To Mix or To Merge: Toward Multi-Domain Reinforcement Learning for Large Language Models}}, 
      author={Haoqing Wang and Xiang Long and Ziheng Li and Yilong Xu and Tingguang Li and Yehui Tang},
      year={2026},
      eprint={2602.12566},
      archivePrefix={arXiv},
      primaryClass={cs.AI},
      url={https://arxiv.org/abs/2602.12566}, 
}

@misc{rw_aiscientistv2,
      title={{The AI Scientist-v2: Workshop-Level Automated Scientific Discovery via Agentic Tree Search}}, 
      author={Yutaro Yamada and Robert Tjarko Lange and Cong Lu and Shengran Hu and Chris Lu and Jakob Foerster and Jeff Clune and David Ha},
      year={2025},
      eprint={2504.08066},
      archivePrefix={arXiv},
      primaryClass={cs.AI},
      url={https://arxiv.org/abs/2504.08066}, 
}

@misc{rw_airesearcher,
      title={{AI-Researcher: Autonomous Scientific Innovation}}, 
      author={Jiabin Tang and Lianghao Xia and Zhonghang Li and Chao Huang},
      year={2025},
      eprint={2505.18705},
      archivePrefix={arXiv},
      primaryClass={cs.AI},
      url={https://arxiv.org/abs/2505.18705}, 
}

@misc{rw_mlestar,
      title={{MLE-STAR: Machine Learning Engineering Agent via Search and Targeted Refinement}}, 
      author={Jaehyun Nam and Jinsung Yoon and Jiefeng Chen and Jinwoo Shin and Sercan {\"O}. Ar{\i}k and Tomas Pfister},
      year={2025},
      eprint={2506.15692},
      archivePrefix={arXiv},
      primaryClass={cs.LG},
      url={https://arxiv.org/abs/2506.15692}, 
}

@misc{rw_posttrainbench,
      title={{PostTrainBench: Can LLM Agents Automate LLM Post-Training?}}, 
      author={Ben Rank and Hardik Bhatnagar and Ameya Prabhu and Shira Eisenberg and Karina Nguyen and Matthias Bethge and Maksym Andriushchenko},
      year={2026},
      eprint={2603.08640},
      archivePrefix={arXiv},
      primaryClass={cs.SE},
      url={https://arxiv.org/abs/2603.08640}, 
}

@misc{rw_ftdojo,
      title={{FT-Dojo: Towards Autonomous LLM Fine-Tuning with Language Agents}}, 
      author={Qizheng Li and Yifei Zhang and Xiao Yang and Xu Yang and Zhuo Wang and Weiqing Liu and Jiang Bian},
      year={2026},
      eprint={2603.01712},
      archivePrefix={arXiv},
      primaryClass={cs.AI},
      url={https://arxiv.org/abs/2603.01712}, 
}

@misc{rw_autotrainess,
      title={{AutoTrainess: Teaching Language Models to Improve Language Models Autonomously}}, 
      author={Zhaojian Yu and Penghao Yin and Shuzheng Gao and Shilin He and Kai Cai and Xiao-Ping Zhang},
      year={2026},
      eprint={2606.31551},
      archivePrefix={arXiv},
      primaryClass={cs.CL},
      url={https://arxiv.org/abs/2606.31551}, 
}

@inproceedings{rw_lamdagent,
      title={{LaMDAgent: An Autonomous Framework for Post-Training Pipeline Optimization via LLM Agents}},
      author={Taro Yano and Yoichi Ishibashi and Masafumi Oyamada},
      booktitle={Proceedings of the 2025 Conference on Empirical Methods in Natural Language Processing},
      year={2025},
      pages={30078--30095},
      publisher={Association for Computational Linguistics},
      doi={10.18653/v1/2025.emnlp-main.1529},
      url={https://aclanthology.org/2025.emnlp-main.1529/},
}

@misc{rw_trex,
      title={{TREX: Automating LLM Fine-tuning via Agent-Driven Tree-based Exploration}},
      author={Zerun Ma and Guoqiang Wang and Xinchen Xie and Yicheng Chen and He Du and Bowen Li and Yanan Sun and Wenran Liu and Kai Chen and Yining Li},
      year={2026},
      eprint={2604.14116},
      archivePrefix={arXiv},
      primaryClass={cs.AI},
      url={https://arxiv.org/abs/2604.14116},
}

@misc{rw_missing_posttraining_ai,
      title={{What is Missing from AI Post-Training AI: An Empirical Analysis}},
      author={Joy Jia Yin Lim and Xin Huang and Hao Peng and Yaxi Lu and Xin Cong and Zhong Zhang and Maosong Sun and Yankai Lin},
      year={2026},
      eprint={2608.19072},
      archivePrefix={arXiv},
      primaryClass={cs.AI},
      url={https://arxiv.org/abs/2608.19072},
}

@misc{rw_llmzero,
      title={{LLMZero: Discovering Adaptive Training Strategies for RL Post-Training via LLM Agents}}, 
      author={Haoyang Fang and Wei Zhu and Boran Han and Alex Zhang and Zhenyu Pan and Shuo Yang and Shuai Zhang and Jiading Gai and Peng Tang and Cuixiong Hu and Xuan Zhu and Huzefa Rangwala and George Karypis and Bernie Wang},
      year={2026},
      eprint={2606.18388},
      archivePrefix={arXiv},
      primaryClass={cs.LG},
      url={https://arxiv.org/abs/2606.18388}, 
}

@misc{rw_pioneer,
      title={{Pioneer Agent: Continual Improvement of Small Language Models in Production}}, 
      author={Dhruv Atreja and Julia White and Nikhil Nayak and Kelton Zhang and Henrijs Princis and George Hurn-Maloney and Ash Lewis and Urchade Zaratiana},
      year={2026},
      eprint={2604.09791},
      archivePrefix={arXiv},
      primaryClass={cs.AI},
      url={https://arxiv.org/abs/2604.09791}, 
}

@misc{rw_datamaster,
      title={{DataMaster: Data-Centric Autonomous AI Research}}, 
      author={Yaxin Du and Xiyuan Yang and Zhifan Zhou and Wanxu Liu and Zixing Lei and Zimeng Chen and Fenyi Liu and Haotian Wu and Yuzhu Cai and Zexi Liu and Xinyu Zhu and WenHao Wang and Linfeng Zhang and Chen Qian and Siheng Chen},
      year={2026},
      eprint={2605.10906},
      archivePrefix={arXiv},
      primaryClass={cs.LG},
      url={https://arxiv.org/abs/2605.10906}, 
}

@misc{rw_andes,
      title={{ANDES: Agent Native Data Evolving Synthesis Tool for Autonomous Instruction Alignment}}, 
      author={Zhengyang Zhao and Shengjie Ye and Lu Ma and Hao Liang and Hengyi Feng and Wentao Zhang},
      year={2026},
      eprint={2606.01279},
      archivePrefix={arXiv},
      primaryClass={cs.AI},
      url={https://arxiv.org/abs/2606.01279}, 
}

@misc{rw_curateevo,
      title={{CurateEvo: Data-Curation Evolving for Agentic Post-Training}}, 
      author={Dingzirui Wang and Xuanliang Zhang and Keyan Xu and Qingfu Zhu and Wanxiang Che},
      year={2026},
      eprint={2607.06140},
      archivePrefix={arXiv},
      primaryClass={cs.CL},
      url={https://arxiv.org/abs/2607.06140}, 
}

@misc{rw_curationbench,
      title={{Can Generalist Agents Automate Data Curation?}}, 
      author={Feiyang Kang and Hanze Li and Adam Nguyen and Mahavir Dabas and Jiaqi W. Ma and Frederic Sala and Dawn Song and Ruoxi Jia},
      year={2026},
      eprint={2606.04261},
      archivePrefix={arXiv},
      primaryClass={cs.AI},
      url={https://arxiv.org/abs/2606.04261}, 
}

@misc{rw_seal,
      title={{Self-Adapting Language Models}}, 
      author={Adam Zweiger and Jyothish Pari and Han Guo and Ekin Akyürek and Yoon Kim and Pulkit Agrawal},
      year={2025},
      eprint={2506.10943},
      archivePrefix={arXiv},
      primaryClass={cs.LG},
      url={https://arxiv.org/abs/2506.10943}, 
}

@misc{rw_agentevolver,
      title={{AgentEvolver: Towards Efficient Self-Evolving Agent System}}, 
      author={Yunpeng Zhai and Shuchang Tao and Cheng Chen and Anni Zou and Ziqian Chen and Qingxu Fu and Shinji Mai and Li Yu and Jiaji Deng and Zouying Cao and Zhaoyang Liu and Bolin Ding and Jingren Zhou},
      year={2025},
      eprint={2511.10395},
      archivePrefix={arXiv},
      primaryClass={cs.LG},
      url={https://arxiv.org/abs/2511.10395}, 
}

@misc{rw_coevolve,
      title={{CoEvolve: Training LLM Agents via Agent-Data Mutual Evolution}}, 
      author={Shidong Yang and Ziyu Ma and Tongwen Huang and Yiming Hu and Yong Wang and Xiangxiang Chu},
      year={2026},
      eprint={2604.15840},
      archivePrefix={arXiv},
      primaryClass={cs.CL},
      url={https://arxiv.org/abs/2604.15840}, 
}

@misc{qwen35,
  author       = {{Qwen Team}},
  title        = {{Qwen3.5-4B}},
  year         = {2026},
  howpublished = {Hugging Face model card},
  url          = {https://huggingface.co/Qwen/Qwen3.5-4B}
}

@misc{qwen35base,
  author       = {{Qwen Team}},
  title        = {{Qwen3.5-4B-Base}},
  year         = {2026},
  howpublished = {Hugging Face model card},
  url          = {https://huggingface.co/Qwen/Qwen3.5-4B-Base}
}

@misc{glm5,
  author        = {{GLM-5 Team}},
  title         = {{GLM-5: From Vibe Coding to Agentic Engineering}},
  year          = {2026},
  eprint        = {2602.15763},
  archivePrefix = {arXiv},
  url           = {https://arxiv.org/abs/2602.15763}
}

@misc{kimi3,
  author        = {{Kimi Team}},
  title         = {{Kimi K3: Open Frontier Intelligence}},
  year          = {2026},
  eprint        = {2607.24653},
  archivePrefix = {arXiv},
  url           = {https://arxiv.org/abs/2607.24653}
}

@misc{mmlupro,
  author        = {Yubo Wang and others},
  title         = {{MMLU-Pro: A More Robust and Challenging Multi-Task Language Understanding Benchmark}},
  year          = {2024},
  eprint        = {2406.01574},
  archivePrefix = {arXiv},
  url           = {https://arxiv.org/abs/2406.01574}
}

@misc{supergpqa,
  author        = {{M-A-P Team}},
  title         = {{SuperGPQA: Scaling LLM Evaluation across 285 Graduate Disciplines}},
  year          = {2025},
  eprint        = {2502.14739},
  archivePrefix = {arXiv},
  url           = {https://arxiv.org/abs/2502.14739}
}

@misc{gpqa,
  author        = {David Rein and others},
  title         = {{GPQA: A Graduate-Level Google-Proof Q\&A Benchmark}},
  year          = {2023},
  eprint        = {2311.12022},
  archivePrefix = {arXiv},
  url           = {https://arxiv.org/abs/2311.12022}
}

@misc{math500,
  author        = {Hunter Lightman and Vineet Kosaraju and Yura Burda and Harri Edwards and Bowen Baker and Teddy Lee and Jan Leike and John Schulman and Ilya Sutskever and Karl Cobbe},
  title         = {{Let's Verify Step by Step}},
  year          = {2023},
  eprint        = {2305.20050},
  archivePrefix = {arXiv},
  url           = {https://arxiv.org/abs/2305.20050}
}

@misc{aime2025,
  author       = {{Mathematical Association of America}},
  title        = {{American Invitational Mathematics Examination 2025}},
  year         = {2025},
  url          = {https://maa.org/maa-invitational-competitions/}
}

@misc{ifeval,
  author        = {Jeffrey Zhou and Tianjian Lu and Swaroop Mishra and Siddhartha Brahma and Sujoy Basu and Yi Luan and Denny Zhou and Le Hou},
  title         = {{Instruction-Following Evaluation for Large Language Models}},
  year          = {2023},
  eprint        = {2311.07911},
  archivePrefix = {arXiv},
  url           = {https://arxiv.org/abs/2311.07911}
}

@inproceedings{ifbench,
  author    = {Valentina Pyatkin and Saumya Malik and Victoria Graf and Hamish Ivison and Shengyi Huang and Pradeep Dasigi and Nathan Lambert and Hannaneh Hajishirzi},
  title     = {{Generalizing Verifiable Instruction Following}},
  booktitle = {Advances in Neural Information Processing Systems},
  year      = {2025},
  url       = {https://arxiv.org/abs/2507.02833}
}

@misc{livecodebench,
  author        = {Naman Jain and King Han and Alex Gu and Wen-Ding Li and Fanjia Yan and Tianjun Zhang and Sida Wang and Armando Solar-Lezama and Koushik Sen and Ion Stoica},
  title         = {{LiveCodeBench: Holistic and Contamination Free Evaluation of Large Language Models for Code}},
  year          = {2024},
  eprint        = {2403.07974},
  archivePrefix = {arXiv},
  url           = {https://arxiv.org/abs/2403.07974}
}

@misc{terminalbench2,
  author        = {Mike A. Merrill and Alexander G. Shaw and Nicholas Carlini and others},
  title         = {{Terminal-Bench: Benchmarking Agents on Hard, Realistic Tasks in Command Line Interfaces}},
  year          = {2026},
  eprint        = {2601.11868},
  archivePrefix = {arXiv},
  url           = {https://arxiv.org/abs/2601.11868}
}

@inproceedings{bfcl,
  author       = {Shishir G. Patil and Huanzhi Mao and Fanjia Yan and Charlie Cheng-Jie Ji and Vishnu Suresh and Ion Stoica and Joseph E. Gonzalez},
  title        = {{The Berkeley Function Calling Leaderboard (BFCL): From Tool Use to Agentic Evaluation of Large Language Models}},
  booktitle    = {Proceedings of the 42nd International Conference on Machine Learning},
  volume       = {267},
  series       = {Proceedings of Machine Learning Research},
  pages        = {48371--48392},
  publisher    = {PMLR},
  year         = {2025},
  url          = {https://proceedings.mlr.press/v267/patil25a.html}
}

@misc{bfcl_v4_scoring,
  author = {Huanzhi Mao and Raymond Tsao and Jingzhuo Zhou and Shishir G. Patil and Joseph E. Gonzalez},
  title = {{BFCL V4: Web Search}},
  year = {2025},
  howpublished = {Berkeley Function Calling Leaderboard official blog},
  url = {https://gorilla.cs.berkeley.edu/blogs/15_bfcl_v4_web_search.html}
}

@misc{tau2bench,
  author        = {Victor Barres and Honghua Dong and Soham Ray and Xujie Si and Karthik Narasimhan},
  title         = {{$\tau^2$-Bench: Evaluating Conversational Agents in a Dual-Control Environment}},
  year          = {2025},
  eprint        = {2506.07982},
  archivePrefix = {arXiv},
  url           = {https://arxiv.org/abs/2506.07982}
}

@misc{codex,
  author       = {{OpenAI}},
  title        = {{Introducing Codex}},
  year         = {2025},
  howpublished = {OpenAI},
  url          = {https://openai.com/index/introducing-codex/}
}

@misc{claudecode,
  author       = {{Anthropic}},
  title        = {{Claude 3.7 Sonnet and Claude Code}},
  year         = {2025},
  howpublished = {Anthropic},
  url          = {https://www.anthropic.com/news/claude-3-7-sonnet}
}

@inproceedings{xia2024less,
  title = {{LESS}: Selecting Influential Data for Targeted Instruction Tuning},
  author = {Xia, Mengzhou and Malladi, Sadhika and Gururangan, Suchin and Arora, Sanjeev and Chen, Danqi},
  booktitle = {Proceedings of the 41st International Conference on Machine Learning},
  series = {Proceedings of Machine Learning Research},
  volume = {235},
  pages = {54104--54132},
  year = {2024},
  url = {https://proceedings.mlr.press/v235/xia24c.html}
}

@inproceedings{pruthi2020tracin,
  title = {Estimating Training Data Influence by Tracing Gradient Descent},
  author = {Pruthi, Garima and Liu, Frederick and Kale, Satyen and Sundararajan, Mukund},
  booktitle = {Advances in Neural Information Processing Systems},
  volume = {33},
  pages = {19920--19930},
  year = {2020},
  url = {https://proceedings.neurips.cc/paper/2020/hash/e6385d39ec9394f2f3a354d9d2b88eec-Abstract.html}
}

@inproceedings{lopezpaz2017gem,
  title = {Gradient Episodic Memory for Continual Learning},
  author = {Lopez-Paz, David and Ranzato, Marc'Aurelio},
  booktitle = {Advances in Neural Information Processing Systems},
  volume = {30},
  year = {2017},
  url = {https://proceedings.neurips.cc/paper_files/paper/2017/hash/f87522788a2be2d171666752f97ddebb-Abstract.html}
}

@inproceedings{schulman2015trpo,
  title = {Trust Region Policy Optimization},
  author = {Schulman, John and Levine, Sergey and Abbeel, Pieter and Jordan, Michael and Moritz, Philipp},
  booktitle = {Proceedings of the 32nd International Conference on Machine Learning},
  series = {Proceedings of Machine Learning Research},
  volume = {37},
  pages = {1889--1897},
  year = {2015},
  url = {https://proceedings.mlr.press/v37/schulman15.html}
}

@inproceedings{simplevqa,
  title = {{SimpleVQA}: Multimodal Factuality Evaluation for Multimodal Large Language Models},
  author = {Cheng, Xianfu and Zhang, Wei and Zhang, Shiwei and Yang, Jian and Guan, Xiangyuan and Wu, Xianjie and Li, Xiang and Zhang, Ge and Liu, Jiaheng and Mai, Yuying and Zeng, Yutao and Wen, Zhoufutu and Jin, Ke and Wang, Baorui and Zhou, Weixiao and Lu, Yunhong and Ji, Hangyuan and Li, Tongliang and Huang, Wenhao and Li, Zhoujun},
  booktitle = {Proceedings of the IEEE/CVF International Conference on Computer Vision},
  pages = {4637--4646},
  year = {2025},
  url = {https://openaccess.thecvf.com/content/ICCV2025/html/Cheng_SimpleVQA_Multimodal_Factuality_Evaluation_for_Multimodal_Large_Language_Models_ICCV_2025_paper.html}
}

@inproceedings{medxpertqa,
  title = {{MedXpertQA}: Benchmarking Expert-Level Medical Reasoning and Understanding},
  author = {Zuo, Yuxin and Qu, Shang and Li, Yifei and Chen, Zhang-Ren and Zhu, Xuekai and Hua, Ermo and Zhang, Kaiyan and Ding, Ning and Zhou, Bowen},
  booktitle = {Proceedings of the 42nd International Conference on Machine Learning},
  series = {Proceedings of Machine Learning Research},
  volume = {267},
  pages = {80961--80990},
  year = {2025},
  url = {https://proceedings.mlr.press/v267/zuo25a.html}
}
}

\clearpage
\onecolumn
\appendix
\begin{center}
{\Large\bfseries ImproveAnyTask: An Autonomous Post-Training Harness for Iterative Model Self-Improvement\par}
\medskip
{\large Supplementary Material\par}
\end{center}
\section{Appendix}

\subsection{Evaluation Splits and Metrics}
\label{app:evaluation-details}

Table~\ref{tab:evaluation-splits} gives the fixed validation and test allocations. We allocate approximately 85\% of each task to testing, rounding the test size down; TAU2 is split and rounded by domain, and BFCL by category. Optimization uses validation feedback only. Scores in the tables and trajectory figures use the held-out test split. The reported scores therefore describe our fixed splits, not the complete public benchmark test sets. Individual scores and reported gains are truncated to two decimal places. Overall means and their gains are computed from displayed scores and rounded to two decimal places; recomputed ablation gains also use displayed scores.

\begin{table}[ht]
\centering
\small
\setlength{\tabcolsep}{5pt}
\caption{Validation and test allocations used in all comparisons, counted as unique examples. For AIME, $\times 5$ denotes five responses per test question. TAU2 is split by domain and BFCL by category.}
\label{tab:evaluation-splits}
\begin{tabular}{@{}lrrl@{}}
\toprule
\textbf{Task} & \textbf{Validation} & \textbf{Test} & \textbf{Primary metric} \\
\midrule
MMLU-Pro & 1,805 & 10,227 & Accuracy \\
SuperGPQA & 3,980 & 22,549 & Accuracy \\
GPQA-Diamond & 30 & 168 & Accuracy \\
MATH-500 & 75 & 425 & Accuracy \\
AIME 2025 & 5 & $25\ (\times 5)$ & Mean accuracy over 125 responses/run \\
IFEval & 82 & 459 & Strict prompt-level accuracy \\
IFBench & 45 & 255 & Prompt-level accuracy \\
LiveCodeBench v6 & 159 & 896 & Pass@1, one response per problem \\
Terminal-Bench 2.0 & 14 & 75 & Task success rate \\
TAU2-Bench & $8+18$ & $42+96$ & Mean of Airline and Telecom \\
BFCL-V4 & 690 & 3,890 & Renormalized no-Web score \\
SimpleVQA & 304 & 1,721 & Correct-answer rate (Opus 4.6 judge) \\
MedXpertQA-MM & 300 & 1,700 & Answer accuracy \\
\bottomrule
\end{tabular}
\end{table}

SimpleVQA~\citep{simplevqa} evaluates short factual answers to image-grounded questions; we use Claude Opus 4.6 to judge correctness. MedXpertQA-MM~\citep{medxpertqa} evaluates expert-level medical questions with images and clinical context, scored by answer accuracy. Both follow the same validation/test isolation and experimental settings as the other tasks.

Because AIME has only 25 test questions, we evaluate each question five times to reduce variability. For each optimization run, accuracy is averaged over all 125 responses; the final result is the mean across two runs, totaling 250 scored responses from the same 25 test questions. This is mean accuracy, not pass@5. Thus, one additional correct response changes a single-run score by 0.8 percentage points and the two-run mean by 0.4 points. LiveCodeBench uses one response per problem. TAU2 equally averages Airline and Telecom scores over 42 and 96 tasks, respectively.

BFCL counts refer to unique scored examples; repeated Memory backends do not create additional test questions. Live Relevance diagnostics are excluded from these counts and from Overall (Appendix~\ref{app:bfcl-scoring}).

\subsection{Implementation and Reproducibility}
\label{app:implementation-details}

\subsubsection{Operational Details}

\noindent\textbf{Error Attribution.}
The harness reuses per-example predictions, scores, and diagnostic records from the current checkpoint. Each candidate problem is associated with affected examples, a proposed cause, and supporting evidence. Causes established by existing evidence are not tested again, and checks that can be completed by reparsing saved outputs require no new generation. Uncertain causes are examined through targeted inference-time interventions or repeated sampling. Recovery is determined by the primary task metric, rather than the disappearance of a surface symptom.

Each problem is initially checked on up to 32 failed examples covering its associated task categories. If the evidence is insufficient to establish priority, the check expands to at most 64 examples. Verified recoveries contribute to $I_T(p)$ with the primary metric's aggregation weights, following Eq.~(\ref{eq:measured-impact}); unchecked recovery indicators remain unknown and are excluded from the calculation. Each round selects one focal problem supported by measured evidence and records the check configuration, example identifiers, and recovered outputs. Interventions also check up to 16 originally correct examples to identify newly introduced errors.

Repeated sampling follows the task's sampling configuration, with at most eight responses per example and optional early stopping after a correct response. Under deterministic decoding, the harness reuses the existing output and applies targeted interventions instead of repeating identical generations.

\noindent\textbf{Checkpoint-Aware Strategy Selection.}
The Update Direction module combines newly researched strategies, historical candidates, and validated assets. It removes duplicates and checks data, code, and resource requirements, typically retaining two or three feasible strategies. Each candidate provides a small set of representative training examples. All candidates are compared using the same checkpoint, fix examples, and keep examples.

The probe computes loss only on target-answer tokens, normalizes by answer length, and then averages across examples. Gradients are restricted to the feed-forward output projection matrices in the last two language decoder layers. A shared random projection produces the fix, keep, and candidate directions. Candidates are ranked by Eq.~(\ref{eq:strategy-score}); ties favor candidates with better-supported reproduction requirements and lower expected execution cost. Selecting a lower-ranked candidate requires a recorded rationale. The harness returns to error attribution only when every candidate has a valid, non-positive score.

The fix direction uses recovered outputs of the focal problem when available, otherwise reference solutions or teacher answers verified by the task evaluator. Keep directions use validation examples that the current checkpoint answers correctly, grouped by the task's existing categories. Missing targets and zero-norm gradients are recorded as unavailable evidence, not zero scores. Validation examples and their recovered outputs are used only for diagnosis and gradient probing; they are excluded from training and replay data. Table~\ref{tab:diagnostic-probe-settings} summarizes the settings. Sample counts are upper bounds: when fewer examples are available, all are used and the actual count is recorded.

\begin{table}[htbp]
\centering
\small
\setlength{\tabcolsep}{6pt}
\renewcommand{\arraystretch}{1.12}
\caption{Settings for diagnostic checks and checkpoint-specific gradient probes.}
\label{tab:diagnostic-probe-settings}
\begin{tabular}{@{}p{0.32\linewidth}p{\dimexpr0.68\linewidth-2\tabcolsep\relax}@{}}
\toprule
\textbf{Setting} & \textbf{Value} \\
\midrule
Failed examples per problem & Up to 32 initially; up to 64 when more evidence is needed \\
Correct examples per intervention & Up to 16 \\
Repeated sampling & Up to 8 responses under stochastic decoding; reuse the existing output under deterministic decoding \\
Candidate strategies per round & Typically 2--3; select 1 \\
Fix examples & Up to 32 \\
Keep examples & Up to 16 per category and 4 categories; use one category when no grouping is available \\
Training examples per candidate & Up to 128 \\
Probe parameters & Feed-forward output projection matrices in the last two language decoder layers \\
Loss & Mean negative log-likelihood over answer tokens; input and padding tokens are masked \\
Gradient projection & 8,192-dimensional CountSketch with a shared mapping across directions \\
Sampling and projection seed & 42; actual example identifiers are also saved \\
Penalty in Eq.~(\ref{eq:strategy-score}) & $\beta=1$ \\
\bottomrule
\end{tabular}
\end{table}

\noindent\textbf{Update Risk and Execution Checks.}
Negative inner products between candidate gradients and gradients of correctly answered examples identify categories at risk of regression. The agent uses this evidence to decide whether to add replay data. Replay examples come from an independent training pool or historical training assets, are checked for correctness by the current model, and have an initial mixing fraction of 10\%. The probe also estimates output sensitivity on correct examples along the candidate update direction. High sensitivity prompts the agent to reduce the proposed learning rate, following Section~\ref{sec:update-direction}.

Before full training, the agent applies the complete data preparation procedure to produce five examples in the final training format. It then runs one optimizer step with the training backend to check data loading, templates, and execution. Failed checks allow up to two correction attempts. This check validates execution rather than comparing candidate accuracy. Its temporary weights are discarded, and full training starts from the current checkpoint after the checks pass.

\subsubsection{Evaluation Configuration and Reproducibility}
\label{app:evaluation-configuration}

\noindent\textbf{Evaluation Settings.}
Temperature, reasoning mode, and generation limits follow explicit benchmark recommendations where available, rather than Qwen's general model-level recommendations. Unspecified settings use the defaults in Table~\ref{tab:evaluation-defaults}, subject to the task-specific settings stated in Section~\ref{sec:experimental-setup}.

\begin{table}[htbp]
\centering
\small
\setlength{\tabcolsep}{12pt}
\caption{Default evaluation settings when a benchmark does not specify the corresponding option.}
\label{tab:evaluation-defaults}
\begin{tabular}{@{}ll@{}}
\toprule
\textbf{Setting} & \textbf{Default} \\
\midrule
Temperature & 0 \\
Thinking mode & Disabled \\
Maximum generation length & 16,384 tokens \\
\bottomrule
\end{tabular}
\end{table}

For each task, the initial model, adapted models, and all baselines share the same data split, input template, inference configuration, answer extraction, and scoring procedure. Diagnostic interventions change only the factor being tested and record deviations from the evaluation configuration; they do not alter formal evaluation conditions. Empty outputs and predictions affected by execution failures are scored as incorrect. Task-specific configurations are saved with the evaluation records and will accompany the evaluation code.

Main and ablation results average two independent complete optimization runs. Response aggregation and task-specific scoring follow Appendix~\ref{app:evaluation-details}. All comparisons measure performance changes relative to the initial checkpoint under the same evaluation protocol.

\noindent\textbf{Run-to-Run Variability.}
Table~\ref{tab:run-std} reports run-to-run variability using the standard deviation (SD) over two independent optimization runs for each setting. MMLU-Pro, SuperGPQA, IFEval, SimpleVQA, and MedXpertQA-MM show relatively low variability, with most standard deviations around 1--2 points. GPQA-Diamond, AIME 2025, Terminal-Bench 2.0, and TAU2-Bench show larger fluctuations. This pattern may partly reflect differences in test-set size and task structure: small test sets make scores more sensitive to individual outcomes, while multi-step execution and tool-use tasks can amplify differences in optimization trajectories and model decisions. Conversely, aggregation over more test examples may help explain the lower variability on the larger knowledge and visual question-answering benchmarks.

\begin{table}[htbp]
\centering
\small
\setlength{\tabcolsep}{4pt}
\renewcommand{\arraystretch}{1.12}
\caption{Run-to-run standard deviation of the results in Table~\ref{tab:main-results}, computed over two independent optimization runs. Values are in percentage points.}
\label{tab:run-std}
\begin{tabular*}{\linewidth}{@{\extracolsep{\fill}}lrrrrrrrr@{}}
\toprule
& \multicolumn{4}{c}{\textbf{Qwen3.5-4B-Base}} & \multicolumn{4}{c}{\textbf{Qwen3.5-4B-Instruct}} \\
\cmidrule(lr){2-5}\cmidrule(l){6-9}
\textbf{Benchmark} & \textbf{Codex} & \textbf{Claude Code} & \textbf{FT-Agent} & \textbf{Ours} & \textbf{Codex} & \textbf{Claude Code} & \textbf{FT-Agent} & \textbf{Ours} \\
\midrule
MMLU-Pro & 1.64 & 1.30 & 1.42 & 1.16 & 1.36 & 1.04 & 1.22 & 0.98 \\
SuperGPQA & 1.40 & 1.85 & 1.05 & 1.23 & 1.10 & 1.42 & 1.23 & 1.62 \\
GPQA-Diamond & 4.21 & 5.05 & 4.21 & 3.37 & 3.37 & 2.53 & 5.89 & 4.21 \\
MATH-500 & 2.66 & 2.99 & 1.66 & 2.33 & 2.00 & 2.33 & 1.50 & 1.33 \\
AIME 2025 & 5.66 & 5.09 & 3.96 & 4.53 & 3.96 & 5.09 & 5.09 & 3.96 \\
IFEval & 1.85 & 1.54 & 1.54 & 1.23 & 1.23 & 1.54 & 0.92 & 1.23 \\
IFBench & 2.77 & 2.22 & 3.88 & 2.77 & 2.22 & 2.77 & 3.60 & 2.22 \\
LiveCodeBench v6 & 4.26 & 3.31 & 3.71 & 2.37 & 3.63 & 2.68 & 3.16 & 2.37 \\
Terminal-Bench 2.0 & 3.77 & 1.89 & 4.71 & 3.77 & 3.77 & 3.77 & 4.71 & 5.66 \\
BFCL-V4$^{\ddagger}$ & 1.89 & 2.34 & 1.48 & 1.76 & 1.40 & 1.67 & 1.51 & 1.22 \\
TAU2-Bench & 5.47 & 6.10 & 4.89 & 4.42 & 4.00 & 5.05 & 3.79 & 3.37 \\
SimpleVQA & 1.81 & 1.31 & 1.48 & 1.11 & 1.31 & 1.68 & 1.19 & 0.99 \\
MedXpertQA-MM & 1.66 & 2.12 & 1.41 & 1.58 & 1.29 & 1.79 & 1.50 & 1.16 \\
\bottomrule
\end{tabular*}
\par\vspace{2pt}
{\footnotesize $^{\ddagger}$BFCL excludes Web Search and renormalizes the remaining weights, following Appendix~\ref{app:bfcl-scoring}.}
\end{table}

\noindent\textbf{Records and Resource Accounting.}
Each run saves per-example predictions and scores, diagnostic evidence, candidate strategies and probe scores, selection rationales, stage durations, GPU-hours, teacher-call costs, and failure counts. Accuracy-based tasks additionally retain raw correct counts and evaluation totals; tasks with grouped aggregation retain per-group counts and weights. Data audits record exact-match counts and handling decisions separately for training data matched against validation and test queries. Test queries remain inaccessible to the optimization agents. Diagnosis, candidate data generation, gradient probing, and execution checks all count toward the end-to-end time budget.

\noindent\textbf{Reproduction Guide.}
We will release the complete harness code, agent prompts and stage definitions, diagnostic and gradient-probing tools, and training and evaluation scripts. The release will include environment dependencies, model versions, split manifests, and task configurations. The reproduction guide will cover environment setup, data acquisition and processing, experiment launch, and result aggregation. A complete execution trace will accompany the release to support inspection of each stage's inputs, decisions, and outputs.

\subsection{Extended IFEval Experiment}
\label{app:ifeval-extended}

This appendix supplements Section~\ref{sec:budget-scaling} with implementation details for the extended Qwen3.5-4B-Base run. C1--C8 follow the chronological order in Figure~\ref{fig:ifeval-extended-budget}. They index candidate checkpoints, not eight accepted updates. This experiment is separate from the Qwen3.5-4B-Instruct ablation in Appendix~\ref{app:ablation-details}.

\noindent\textbf{Data Sources and Preparation.}
The initial mixture draws on \href{https://huggingface.co/datasets/allenai/tulu-3-sft-personas-instruction-following}{Tulu~3 Persona-IF}, \href{https://huggingface.co/datasets/sojuL/RubricHub_v1}{RubricHub's released best-of-six responses}, and \href{https://huggingface.co/datasets/nvidia/Nemotron-Instruction-Following-Chat-v1}{Nemotron Instruction-Following Chat v1}. We use the selected RubricHub responses, not its intermediate candidate collection, and a filtered subset of Nemotron rather than its full post-training recipe. Later trials construct targeted examples and retain answers that pass the official executable instruction checks. These checks validate the specified constraints; they do not independently establish factual correctness or the quality of a reasoning trace.

Table~\ref{tab:ifeval-data-details} reports the final exported training-set sizes. Component counts refer to collections before final selection, deduplication, and overlap filtering, and therefore need not sum to the exported totals. C5 starts from C3 rather than the unsuccessful C4 checkpoint. It revises C4's data strategy while keeping the earlier model.

\begingroup
\small
\setlength{\tabcolsep}{4pt}
\renewcommand{\arraystretch}{1.13}
\begin{longtable}{@{}p{0.055\linewidth}p{0.09\linewidth}p{\dimexpr0.745\linewidth-6\tabcolsep\relax}>{\raggedleft\arraybackslash}p{0.11\linewidth}@{}}
\caption{Data preparation and training ancestry for the extended IFEval Base experiment. Counts in the last column are final exported examples.}\label{tab:ifeval-data-details}\\
\toprule
\textbf{Ckpt.} & \textbf{Parent} & \textbf{Strategy and data construction} & \textbf{Examples} \\
\midrule
\endfirsthead
\multicolumn{4}{l}{\tablename~\thetable\ (continued)}\\
\toprule
\textbf{Ckpt.} & \textbf{Parent} & \textbf{Strategy and data construction} & \textbf{Examples} \\
\midrule
\endhead
\midrule
\multicolumn{4}{r}{Continued on next page}\\
\endfoot
\bottomrule
\endlastfoot
C1 & Initial Base & \textbf{Persona-conditioned constraint tuning.} Use released instruction--response pairs covering keyword, formatting, length, and capitalization constraints. Preserve message structure, validate formatting, deduplicate, and audit exact overlap with evaluation queries. & 29,900 \\
\addlinespace
C2 & C1 & \textbf{Rubric-guided response tuning.} Combine persona-conditioned examples with approximately 26K released responses selected by best-of-six rubric scoring. Convert the selected pairs to the training message format before merging and filtering. & 55,998 \\
\addlinespace
C3 & C2 & \textbf{Multi-source instruction mixture tuning.} Combine approximately 30K persona-conditioned, 24K rubric-selected, and 16K Nemotron examples. Retain non-thinking responses from the additional source, remove visual markers, and filter by length. & 69,900 \\
\addlinespace
C4 & C3 & \textbf{Execution-verified targeted constraint tuning.} Mix approximately 23.4K verified synthetic examples with 13K replay examples. Target character and keyword frequencies, punctuation, capitalization, and sentence counts identified by the preceding error analysis. & 36,299 \\
\addlinespace
C5 & C3 & \textbf{Replay-preserving additive constraint tuning.} Retain the full 69,900-example C3 mixture and add approximately 23,750 execution-verified examples, then deduplicate and filter. Unlike C4, this trial adds targeted supervision without replacing the broader mixture. & 93,000 \\
\addlinespace
C6 & C5 & \textbf{Residual-error-guided constraint augmentation.} Extend the 93K-example mixture with 14,524 verified examples covering remaining lexical, counting, formatting, length, and punctuation errors. & 107,000 \\
\addlinespace
C7 & C6 & \textbf{Explicit constraint reasoning with self-verification.} Combine 2,575 Qwen3.5-122B-generated reasoning traces with 18K verified answers augmented by rationales based on computable constraint checks. Retain examples whose final answers pass the checker. & 20,000 \\
\addlinespace
C8 & C7 & \textbf{Precision-focused self-verification refinement.} Extend C7's 20,575-example pre-filter collection with 562 verified teacher-generated traces and 8K additional self-checking examples for exact-counting and lexical constraints, then filter the combined collection. & 24,500 \\
\end{longtable}
\endgroup

\noindent\textbf{Training Configuration.}
All eight trials use full-parameter post-training for three epochs with a global batch size of 128. Table~\ref{tab:ifeval-training-details} lists the learning rates and sequence limits. The training sequence limit is distinct from the evaluation context window and maximum output length. The augmented self-checking examples should not be interpreted as independently sampled teacher reasoning traces.

\begin{table}[ht]
\centering
\small
\setlength{\tabcolsep}{7pt}
\renewcommand{\arraystretch}{1.13}
\caption{Training settings for the extended IFEval run. The sequence limit is measured in tokens.}
\label{tab:ifeval-training-details}
\begin{tabular}{@{}lrr@{}}
\toprule
\textbf{Checkpoints} & \textbf{Learning rate} & \textbf{Sequence limit} \\
\midrule
C1--C3 & $1\times10^{-5}$ & 4,096 \\
C4 & $5\times10^{-6}$ & 4,096 \\
C5--C6 & $1\times10^{-5}$ & 4,096 \\
C7--C8 & $1\times10^{-5}$ & 8,192 \\
\bottomrule
\end{tabular}
\end{table}

\subsubsection{Evaluation and Resource Accounting}
\label{app:ifeval-evaluation}

\noindent\textbf{Generation and Scoring.}
The evaluation prompts and official strict prompt-level scoring procedure remain fixed. All checkpoints use the same IFEval evaluation configuration as the main experiments (Section~\ref{sec:experimental-setup}). C7--C8 use reasoning-augmented training data but retain these evaluation settings.

\noindent\textbf{Resource Budget.}
The extended experiment increases the time limit to 100 hours under the same resource allocation and timing rules as the main experiments, with the three-round limit removed. Checkpoints are equally spaced; their labels give elapsed time. Chronological order differs from training ancestry: C4 is an executed attempt even though C5 resumes from C3.

\subsection{Additional Ablation Studies}
\label{app:ablation-details}

This appendix supplements Section~\ref{sec:ablation} with fine-grained mechanism ablations, module-level replacement definitions, and recorded optimization paths, including an additional MedXpertQA-MM case study. All runs start from Qwen3.5-4B-Instruct and follow the resource limits and end-to-end timing protocol in Section~\ref{sec:experimental-setup}, stopping after 24 hours or three optimization rounds, whichever occurs first. The data sizes and training configurations in Appendix~\ref{app:ifeval-extended} belong to the separate Base experiment and are not reused as descriptions of these ablations.

\noindent\textbf{Key-Mechanism Ablations.}
Beyond replacing entire modules, we further ablate key mechanisms within EA and UD to assess recovery-based problem prioritization and checkpoint-aware strategy selection. We compare three variants while retaining the remaining harness procedures: \emph{EA: error-frequency ranking} selects the focal problem by its associated error count instead of verified recovery impact; \emph{UD: w/o $S$-guided selection} lets the agent select among feasible strategies without using $S(s)$ to guide the choice; and \emph{UD: fix-only score} sets $\beta=0$ in Eq.~\ref{eq:strategy-score}, retaining fix alignment while removing the conflict penalty. Full Harness, w/o EA, and w/o UD are included from Table~\ref{tab:ablation-results} for comparison.

\begin{table}[ht]
\centering
\small
\setlength{\tabcolsep}{3pt}
\renewcommand{\arraystretch}{1.15}
\caption{Ablations of key mechanisms within EA and UD on Qwen3.5-4B-Instruct. Scores are percentages; Avg. is the unweighted mean across the six tasks. Bold marks the best result in each column.}
\label{tab:mechanism-ablation}
\begin{tabular*}{\linewidth}{@{\extracolsep{\fill}}lrrrrrrr@{}}
\toprule
\textbf{Setting} & \textbf{MMLU-Pro} & \textbf{MATH-500} & \textbf{LCB v6} & \textbf{TAU2} & \textbf{IFEval} & \textbf{SimpleVQA} & \textbf{Avg.} \\
\midrule
\rowcolor{oursblue}
Full Harness & \textbf{69.33} & \textbf{63.52} & \textbf{40.62} & \textbf{44.94} & 83.66 & \textbf{51.88} & \textbf{58.99} \\
w/o EA & 61.20 & 56.00 & 25.16 & 39.73 & 82.35 & 46.80 & 51.87 \\
w/o UD & 60.30 & 56.70 & 32.81 & 41.66 & 83.22 & 48.92 & 53.94 \\
\midrule
EA: error-frequency ranking & 66.53 & 60.82 & 36.71 & 43.75 & 83.44 & 50.23 & 56.91 \\
UD: w/o $S$-guided selection & 64.92 & 59.76 & 35.93 & 43.15 & 83.33 & 50.14 & 56.21 \\
UD: fix-only score ($\beta=0$) & 67.87 & 62.11 & 38.95 & 44.41 & \textbf{83.76} & 51.10 & 58.03 \\
\bottomrule
\end{tabular*}
\end{table}

The full harness achieves the highest average and leads on five of the six tasks (Table~\ref{tab:mechanism-ablation}). Ranking problems by error frequency reduces the average from 58.99 to 56.91, supporting the value of verified recovery evidence over error prevalence alone. Removing $S$-guided selection lowers it to 56.21, showing the benefit of checkpoint-specific evidence when choosing among researched strategies. Using fix alignment alone scores 58.03, 0.96 points below the full method, supporting the contribution of the conflict penalty. On IFEval, the fix-only variant is 0.10 points higher; this small difference may reflect stochastic variation in LLM-driven optimization. The full method nevertheless provides the strongest aggregate performance. Each targeted ablation also remains above its corresponding whole-module removal, indicating that these mechanisms contribute alongside the broader EA and UD procedures.

\noindent\textbf{Module Replacements.}
Table~\ref{tab:ablation-replacements} groups the procedures in Sections~\ref{sec:error-attribution}--\ref{sec:executable-updates} by the module being replaced. The underlying agent retains access to tools, artifact inspection, and ordinary debugging in every variant. In particular, w/o ER retains the candidate samples and gradient probes used by UD, while omitting the execution checks applied before full training. The replacements therefore assess the combined effect of each prescribed module, including its influence on subsequent decisions.

\begin{table}[ht]
\centering
\small
\setlength{\tabcolsep}{4pt}
\renewcommand{\arraystretch}{1.15}
\caption{Module-level replacements aligned with Methods. ER denotes Execution-Guided Refinement within Executable Model Updates.}
\label{tab:ablation-replacements}
\begin{tabular}{@{}p{0.12\linewidth}p{\dimexpr0.44\linewidth-2\tabcolsep\relax}p{\dimexpr0.44\linewidth-2\tabcolsep\relax}@{}}
\toprule
\textbf{Variant} & \textbf{Procedure omitted} & \textbf{Basic alternative retained} \\
\midrule
w/o EA & Joint metric- and case-level analysis; conditional cause verification through inference-time interventions or repeated sampling; recovery-based prioritization and focal-problem selection. & An unstructured summary of evaluation feedback guides the next action. General feedback analysis remains available. \\
\addlinespace
w/o UD & Structured candidate comparison, reproduction-difficulty and resource checks, and outcome-aware reuse; gradient alignment and regression-risk probes; probe-guided selection, replay, learning-rate adjustment, and return to problem selection when no candidate has positive alignment. & Ordinary research and agent judgment select a feasible strategy and training settings. Resource inspection and later reconsideration remain possible. \\
\addlinespace
w/o ER & Prescribed sample, data, and template checks; short training validation; iterative correction of data and training pipelines before full execution. & Ordinary model updates and debugging. Candidate sampling and gradient probes used by UD remain available; execution issues may be repaired when encountered. \\
\bottomrule
\end{tabular}
\end{table}

\noindent\textbf{Recorded Optimization Decisions.}
Table~\ref{tab:ablation-paths} summarizes strategy and execution records for MMLU-Pro, MATH-500, LiveCodeBench v6, TAU2-Bench, and IFEval, together with an additional MedXpertQA-MM case study. The MedXpertQA-MM records illustrate optimization behavior and are not included in the six-task averages in Table~\ref{tab:ablation-results}. Steps denote successive recorded decisions, which can include data preparation, training, adapter rescaling, or implementation repair; they do not necessarily denote three newly trained checkpoints. Data counts are included where recorded and are specific to the model initialization used in that run.

\begingroup
\small
\setlength{\tabcolsep}{4pt}
\renewcommand{\arraystretch}{1.16}
\begin{longtable}{@{}p{0.115\linewidth}p{\dimexpr0.295\linewidth-2\tabcolsep\relax}p{\dimexpr0.295\linewidth-2\tabcolsep\relax}p{\dimexpr0.295\linewidth-2\tabcolsep\relax}@{}}
\caption{Recorded strategy and execution paths for four Instruct-model variants on six tasks, including an additional MedXpertQA-MM case study. Steps index recorded decisions and may include training or corrective adjustments.}\label{tab:ablation-paths}\\
\toprule
\textbf{Variant} & \textbf{Step 1} & \textbf{Step 2} & \textbf{Step 3} \\
\midrule
\endfirsthead
\multicolumn{4}{l}{\tablename~\thetable\ (continued)}\\
\toprule
\textbf{Variant} & \textbf{Step 1} & \textbf{Step 2} & \textbf{Step 3} \\
\midrule
\endhead
\midrule
\multicolumn{4}{r}{Continued on next page}\\
\endfoot
\bottomrule
\endlastfoot
\multicolumn{4}{@{}l}{\textbf{MMLU-Pro}}\\*
Full & Train on 20K long mathematical reasoning examples. & Train on 20K concise final-answer examples; repair end-of-sequence handling and premature termination. & Introduce cross-domain supervision using complete, verified teacher responses. \\
w/o EA & Train on 10K auxiliary knowledge questions with short rationales. & Expand to 18K examples and add output-format anchors. & Introduce structured short-form reasoning supervision. \\
w/o UD & Apply critique-guided distillation. & Switch to long-form mathematical reasoning data. & Introduce truth-teller/liar reasoning exercises. \\
w/o ER & Directly train on long mathematical reasoning traces. & Switch to final-answer supervision while termination and template issues remain unresolved. & Train on 215K examples without checking teacher-response completeness. \\
\midrule
\multicolumn{4}{@{}l}{\textbf{MATH-500}}\\*
Full & Train on 20K mathematical examples with final-answer supervision. & Incorporate 20K examples with extended reasoning. & Select 4,947 high-quality reasoning trajectories for further training. \\
w/o EA & Apply LoRA training on a general mathematical problem set. & Set the adapter scaling factor to 0.48. & Set the adapter scaling factor to 0.30. \\
w/o UD & Adopt final-answer training data. & Switch to a large mathematical instruction collection. & Return to long-form mathematical reasoning data. \\
w/o ER & Train on final-answer examples with limited configuration checks. & Apply a stronger LoRA update. & Reduce update strength as a corrective adjustment. \\
\midrule
\multicolumn{4}{@{}l}{\textbf{LiveCodeBench v6}}\\*
Full & Train on concise code solutions validated through execution. & Distill low-temperature reasoning responses selected for correctness. & Add higher-temperature, more diverse responses and replay data. \\
w/o EA & Apply LoRA training on concise code solutions. & Expand the same type of short-solution data. & Adjust sequence length and adapter strength. \\
w/o UD & Begin with general code instruction tuning. & Switch to difficult code-reasoning examples. & Mix code-reasoning examples with general code replay. \\
w/o ER & Directly use code examples with long reasoning traces. & Proceed with unresolved thinking-template and termination checks. & Partially repair generation and execution settings in the training-data pipeline. \\
\midrule
\multicolumn{4}{@{}l}{\textbf{TAU2-Bench}}\\*
Full & Train on 12K verifiable tool-use trajectories. & Introduce airline trajectories emphasizing evidence-supported write operations. & Repair the training-data user simulator and add 680 successful telecom trajectories. \\
w/o EA & Train on general verifiable tool-use trajectories. & Continue with airline-domain data. & Add retail-domain trajectories. \\
w/o UD & Begin with basic tool-trajectory training. & Generate 88 telecom-domain trajectories. & Switch to airline write-operation trajectories. \\
w/o ER & Select telecom-domain trajectory generation. & Encounter a training-data simulator unable to invoke user-side tools. & Partially repair deduplication and schema issues. \\
\midrule
\multicolumn{4}{@{}l}{\textbf{IFEval}}\\*
Full & Train on persona-conditioned requests with verifiable constraints. & Add rubric-selected best-of-six responses. & Expand to a three-source mixture with conversational instruction data. \\
w/o EA & Begin with persona-conditioned instruction tuning. & Expand general instruction-following data. & Continue mixing general instruction data. \\
w/o UD & Begin with targeted constraint examples. & Switch to a general instruction-following mixture. & Adopt an executable data-synthesis strategy without structured comparison. \\
w/o ER & Generate targeted constraint examples. & Replace the broader mixture with specialized data. & Continue with limited replay coverage and small-scale validation. \\
\midrule
\multicolumn{4}{@{}l}{\textbf{MedXpertQA-MM}}\\*
Full & Compare reported gains and reproduction requirements for MedEvo, m23k RLVR, and MedSyn SFT; prioritize MedSyn image--text SFT with available resources. & Try PMC-VQA reasoning supervision; use error cases and answer-extraction rates to identify capability and format regressions. & Switch to concise two-turn supervision, a lower learning rate, and WiSE-FT interpolation with a fixed coefficient; re-evaluate and retain the best checkpoint. \\*
w/o EA & Apply SFT to general medical image--text data, using overall accuracy to summarize adaptation needs. & Add PMC-VQA and MedMCQA to broaden medical knowledge and reasoning coverage. & Continue adjusting data-mixture proportions and update strength, with limited differentiation of supervision for visual understanding, clinical reasoning, and output formatting. \\*
w/o UD & Prioritize MedEvo-style error-trajectory repair based on its reported gains on the target benchmark. & Build a pipeline using public training questions to generate student reasoning, localize errors, and produce local continuations; determine resource requirements during implementation. & Scale down trajectory repair for SFT based on generation cost and usable sample counts. \\*
w/o ER & Perform medical image--text SFT after routine data conversion, omitting prescribed sample checks and short training validation. & Introduce long reasoning supervision; inspect rationale--answer consistency, two-turn templates, and answer extraction after full training. & Use execution feedback to repair formatting, shorten supervision, and reduce update strength; retrain and evaluate with the remaining budget. \\
\end{longtable}
\endgroup

\begin{figure}[!ht]
    \centering
    \includegraphics[width=\textwidth]{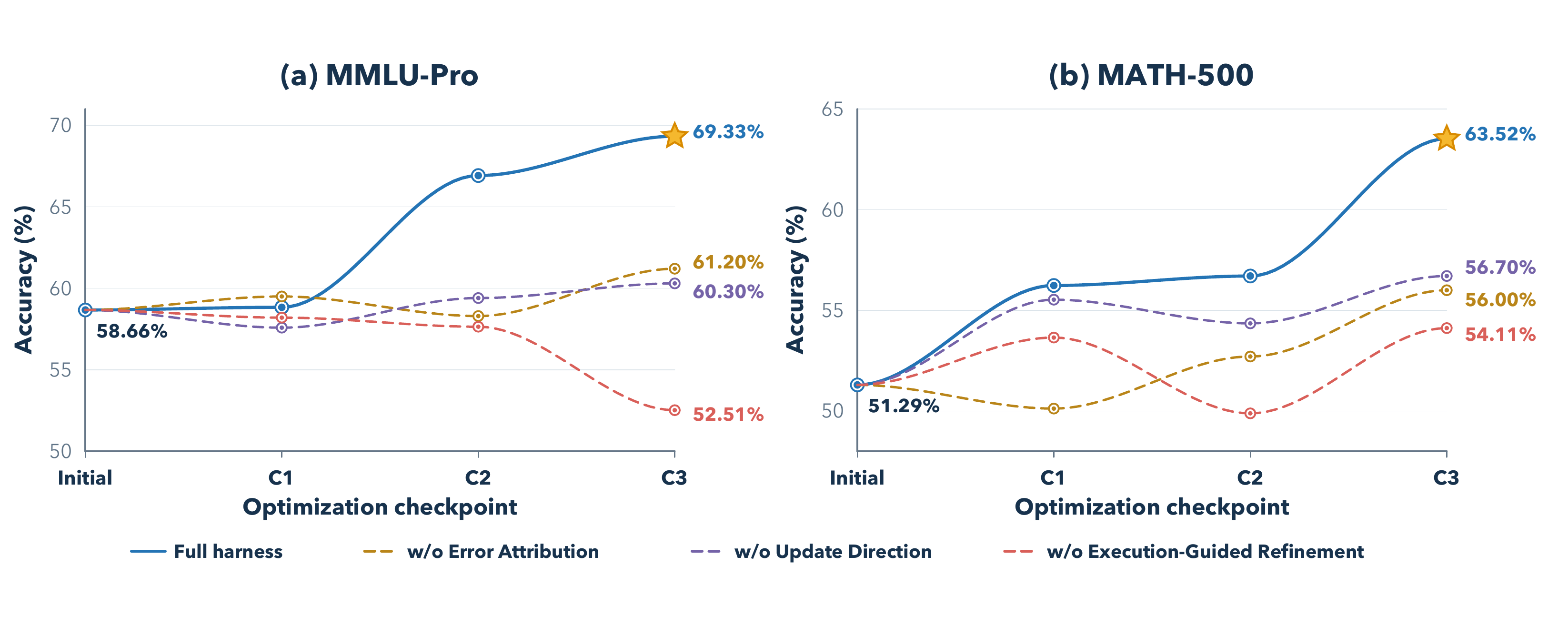}
    \caption{Optimization trajectories for Qwen3.5-4B-Instruct on MMLU-Pro and MATH-500. Initial and final scores match Table~\ref{tab:ablation-results}. Curves connect measured checkpoints for visualization.}
    \label{fig:ablation-trajectories}
\end{figure}

\noindent\textbf{Optimization Trajectories.}
Figure~\ref{fig:ablation-trajectories} compares the MMLU-Pro and MATH-500 trajectories. The full harness improves across successive checkpoints, whereas the replacement variants fluctuate and finish below it. Without Execution-Guided Refinement, MMLU-Pro ends below its initial score, while MATH-500 partially recovers after a decline.

\noindent\textbf{Optimization Focus and Strategy Selection.}
The MATH-500 run without EA moves from general mathematical training to two adapter-scaling adjustments, whereas the full harness changes the supervision and then selects reasoning trajectories. The variant without UD moves between final-answer supervision, a large instruction collection, and long-form reasoning. These decisions illustrate different uses of the available budget. A strategy change is not itself a failure; the comparison concerns how each run diagnoses its needs and selects the next action.

\noindent\textbf{Execution Checks and Recovery.}
The variants without ER still make corrective adjustments, including adapter rescaling, configuration changes, and schema repairs. Their records also contain unresolved termination behavior, unchecked response completeness, and simulator-tool failures. The LiveCodeBench configuration changes and TAU2 simulator repairs concern training-data generation and execution debugging; all models use the same fixed final evaluation environment. Our mechanism organizes checks before full execution, rather than introducing debugging as a new capability. The records help interpret the observed outcomes, but do not establish that any one implementation issue alone caused a particular score change.

\subsection{BFCL Scoring Without Web Search}
\label{app:bfcl-scoring}

The official BFCL-V4 score assigns 40\% to Agentic, which equally averages Web Search and Memory~\citep{bfcl_v4_scoring}. Counting omitted Web Search as zero limits Overall to 80\%. For our no-Web metric, we remove its 20\% weight and renormalize the remaining components:
\[
J_{\mathrm{no\text{-}web}} =
\frac{0.1J_{\mathrm{non\text{-}live}}+0.1J_{\mathrm{live}}+0.1J_{\mathrm{irr}}
+0.3J_{\mathrm{multi}}+0.2J_{\mathrm{memory}}}{0.8}.
\]
This gives weights of 12.5\% each to Non-Live, Live, and Irrelevance, 37.5\% to Multi-Turn, and 25\% to Memory. We retain the original within-section aggregation, including equal averaging across the three Memory backends. This subset metric is not directly comparable to the official full-task Overall score.

We allocate 85\% of each lowest-level category to held-out testing, with sample counts rounded down. The split contains 3,890 scored unique examples and 4,152 scored executions, since the same 131 Memory questions use three backends. Another 13 Live Relevance examples are diagnostic and excluded from Overall. Category-wise rounding can yield a different total from rounding the combined dataset. Shared Memory question IDs are assigned to a split before backend execution.

\clearpage
\subsection{Practical Value and Implications}
\label{app:practical-value}

Task-specific optimization typically requires practitioners to inspect errors, review relevant methods, prepare data, and revise training configurations over repeated trials. ImproveAnyTask automates these activities and reduces reliance on manual decisions at each iteration. Its verification and selection procedures assess uncertain diagnoses and candidate strategies before committing to full training, helping direct a limited budget toward promising updates. Our experiments evaluate the resulting performance under a matched compute budget; reductions in human working time remain to be quantified.

Developing a general-purpose Instruct model requires broad instruction-data curation and post-training. For users targeting a specific task, our results demonstrate a more focused route: starting from an available Base checkpoint, the harness can automatically reach or exceed the original Instruct model's performance on some tasks within 24 hours using resources equivalent to eight H20 GPUs. For example, on MATH-500 and LiveCodeBench v6, the adapted Base model scores 71.52 and 61.60, compared with 51.29 and 11.49 for the original Instruct model (Table~\ref{tab:main-results}). These results concern task-specific adaptation and do not establish equivalence in general instruction following or alignment.

The optimization process also produces reusable evidence and artifacts. Retained diagnoses, strategy evaluations, data-generation scripts, and execution records help subsequent iterations build on previous attempts and make optimization decisions easier to inspect. This is useful when the model's remaining weaknesses change after each update, requiring continued adaptation beyond a single fixed training recipe.

\subsection{Limitations and Future Work}
\label{app:limitations}

Following prior work on task-specific post-training~\citep{rw_posttrainbench,rw_ftdojo}, we focus on target-task performance. We do not systematically measure whether specialization affects general capabilities, given the substantial evaluation effort and the many combinations of tasks, models, and training strategies. The substantial task-specific gains support applications that need focused model adaptation without repeated manual experimentation. Future work will study how to preserve general capabilities during this process.

We also restrict executable updates to supervised post-training. Extending the harness to reinforcement learning and broader post-training objectives is left to future work.

\section{Ethics and Reproducibility}
\subsection*{Ethics Statement}

ImproveAnyTask uses released resources and model-generated data for task-specific post-training. These sources may contain biases, factual errors, or sensitive information, and higher task scores do not establish safety or fairness. Validation feedback is isolated from final testing. Exact-overlap auditing does not rule out semantic contamination in external data, and repeated use of validation feedback can still lead to validation-set overfitting. Autonomous code execution and external API calls also introduce privacy, security, and resource-use risks. Practical deployment therefore requires appropriate access controls and review of generated code and data. Third-party models, code, and datasets remain subject to their licenses and usage conditions.

\subsection*{Reproducibility Statement}

Section~\ref{sec:methods} describes the optimization workflow, and Section~\ref{sec:experimental-setup} specifies the models, tasks, evaluation protocol, and resource limits. The user-facing task prompt is, for example, ``Autonomously optimize the model for MMLU-Pro.'' This request initiates the harness workflow described in Section~\ref{sec:methods}. Appendix~\ref{app:ifeval-extended} provides checkpoint ancestry, data construction, training-set sizes, and training and generation settings for the extended IFEval run. Appendix~\ref{app:ablation-details} describes the baseline replacements used in the ablations and the recorded strategies and execution issues for each task. We will release the complete source code and execution traces upon acceptance to support reproduction. Although we set the optimization agent's API temperature to $0$, external resource availability, API behavior, and training randomness can still affect the full search trajectory.

\end{document}